\documentclass[runningheads]{llncs}
\usepackage{graphicx}

\usepackage{tikz}
\usepackage{orcidlink} 
\usepackage{comment}
\usepackage{amsmath,amssymb} 
\usepackage{color}
\usepackage{subcaption}
\usepackage{amsmath}
\usepackage{amssymb}
\usepackage{booktabs}
\usepackage{multirow}
\usepackage{hyperref}
\usepackage[font=small]{caption}
\usepackage[accsupp]{axessibility}  

\usepackage[capitalize]{cleveref}
\crefname{section}{Sec.}{Secs.}
\Crefname{section}{Section}{Sections}
\Crefname{table}{Table}{Tables}
\crefname{table}{Tab.}{Tabs.}

\begin{document}
\pagestyle{headings}
\mainmatter
\def\ECCVSubNumber{3904}  

\title{CANF-VC: Conditional Augmented Normalizing Flows for Video Compression} 

\titlerunning{CANF-VC}
%
\author{Yung-Han Ho\inst{1}\orcidlink{0000-0001-9384-3258} \and
Chih-Peng Chang\inst{1}\orcidlink{} \and
Peng-Yu Chen\inst{1}\orcidlink{0000-0002-9016-3297} \and
Alessandro Gnutti\inst{2}\orcidlink{0000-0002-8308-0776} \and
Wen-Hsiao Peng\inst{1}\orcidlink{0000-0002-4421-8031}}
\authorrunning{Ho et al.}
%
\institute{Department of Computer Science, National Yang Ming Chiao Tung University, Taiwan \email{wpeng@cs.nctu.edu.tw} \and Department of Information Engineering, CNIT - University of Brescia, Italy
\email{alessandro.gnutti@unibs.it}}

\maketitle

\begin{abstract}
This paper presents an end-to-end learning-based video compression system, termed CANF-VC, based on conditional augmented normalizing flows (CANF). Most learned video compression systems adopt the same hybrid-based coding architecture as the traditional codecs. Recent research on conditional coding has shown the sub-optimality of the hybrid-based coding and opens up opportunities for deep generative models to take a key role in creating new coding frameworks. CANF-VC represents a new attempt that leverages the conditional ANF to learn a video generative model for conditional inter-frame coding. We choose ANF because it is a special type of generative model, which includes variational autoencoder as a special case and is able to achieve better expressiveness. CANF-VC also extends the idea of conditional coding to motion coding, forming a purely conditional coding framework. Extensive  experimental results on commonly used datasets confirm the superiority of CANF-VC to the state-of-the-art methods. The source code of CANF-VC is available at \url{https://github.com/NYCU-MAPL/CANF-VC}.
\end{abstract}

\section{Introduction}
Video compression is an active research area. The video traffic continues to grow exponentially due to an increased demand for various emerging video applications, particularly on social media platforms and mobile devices. The traditional video codecs, such as HEVC~\cite{hevc} and VVC~\cite{vvc}, are still thriving towards being more efficient, hardware-friendly, and versatile. However, their backbones follow the hybrid-based coding framework--namely, spatial/temporal predictive coding plus transform-based residual coding--which has not changed since decades ago.   

The arrival of deep learning spurs a new wave of developments in end-to-end learned image and video compression~\cite{minnen2018,cheng2020,iwave,fvc,dcvc,elfvc}. The seminal work~\cite{googleiclr18} by Ball{\'e} \emph{et al.} connects for the first time the learning of an image compression system to learning a variational generative model, known as the variational autoencoder (VAE)~\cite{vae}. VAE involves learning the autoencoder network jointly with the prior distribution network by maximizing the variational lower bound (ELBO) on the image likelihood $p(x)$. Many follow-up works have been centered around enhancing the autoencoder network~\cite{cheng2020,nlaic} and/or improving the prior modeling~\cite{minnen2018,cheng2020}. Lately, there have been few attempts at introducing normalizing flow models~\cite{iwave,anfic} to learned image compression


 
Inspired by the success of learned image compression, research on learned video compression is catching up quickly. However, most end-to-end learned video compression systems~\cite{dvclu,dvcpro,mlvc,rafc,elfvc} were developed based primarily on the traditional, hybrid-based coding architecture, replacing key components, such as inter-frame prediction and residual coding, with neural networks. The idea of residual coding is to encode a target frame $x_t$ by coding the prediction residual $r_t = x_t-x_c$ between $x_t$ and its motion-compensated reference frame $x_c$. The recent revisit of residual coding as a problem of conditional coding in~\cite{mmsp,iclrw,dcvc} opens up a new dimension of thinking. Arguably, the entropy $H(x_t-x_c)$ of the residual between the coding frame $x_t$ and its motion-compensated reference frame $x_c$ is greater than or equal to the conditional entropy $H(x_t|x_c)$, i.e.~$H(x_t-x_c) \geq H(x_t|x_c)$. How to learn $p(x_t|x_c)$ is apparently the key to the success of conditional coding.
 
In this paper, we present a conditional augmented normalizing flow-based video compression (CANF-VC) system, which is inspired partly by the ANF-based image compression (ANFIC)~\cite{anfic}. However, while ANFIC~\cite{anfic} adopts ANF to learn the (unconditional) image distribution $p(x)$ for image compression, we address video compression from the perspective of learning a video generative model by maximizing the conditional likelihood $p(x_t|x_c)$. We choose the conditional augmented normalizing flow (CANF) to learn $p(x_t|x_c)$, because ANF is a special type of generative model, which includes VAE as a special case and is able to achieve superior expressiveness to VAE. 

Our work has three main contributions: (1) CANF-VC is the first normalizing flow-based video compression system that leverages CANF to learn a video generative model for conditional inter-frame coding;
(2) CANF-VC extends the idea of conditional inter-frame coding to conditional motion coding, forming a purely conditional coding framework; and (3) extensive experimental results confirm the superiority of CANF-VC to the state-of-the-art methods. 

\section{Related Work}
\label{sec:related}

\vspace{-0.4em}
\subsection{Learned Video Compression}
\vspace{-0.4em}
End-to-end learned video compression is a hot research area. DVC~\cite{dvclu} presents the first end-to-end learned video coding framework based on temporal predictive coding. Since then, there have been several improvements on learning-based motion-compensated prediction. Agustsson \emph{et al.}~\cite{ssf} estimate the uncertainty about the flow map in forming a frame predictor, with a scale index sent for each pixel to determine a spatially-varying Gaussian kernel for blurring the reference frame. Liu \emph{et al.}~\cite{nvc} perform feature-domain warping in a coarse-to-fine manner. Hu \emph{et al.}~\cite{fvc} adopt deformable convolution for feature warping. Lin \emph{et al.}~\cite{mlvc} and Yang \emph{et al.}~\cite{hlvc} form a multi-hypothesis prediction from multiple reference frames. 
To reduce motion overhead, Lin \emph{et al.}~\cite{mlvc} use predictive motion coding by extrapolating a flow map predictor from the decoded flow maps. Rippel \emph{et al.}~\cite{elfvc} use the flow map predictor for motion compensation and signal an incremental flow map between the resulting motion-compensated frame and the target frame. Hu \emph{et al.}~\cite{rafc} adapt, either locally or globally, the resolution of the flow map features. Most learned video codecs encode the residual frame or the residual flow map by a variational autoencoder (VAE)-based image coder~\cite{googleiclr18}. Some additionally leverage a recurrent neural network to propagate causal, temporal information in forming a temporal prior for entropy coding~\cite{rlvc,nvc}.
 
\vspace{-0.4em}
\subsection{Conditional Coding}
\vspace{-0.2em}
\label{subsec:CondCoding}
The idea of encoding the residual signal has recently been revisited from the information-theoretic perspective. Ladune \emph{et al.} \cite{mmsp}  show that coding a video frame $x_t$ conditionally based on its motion-compensated reference frame $x_c$ can achieve a lower entropy rate than coding the residual signal $x_t - x_c$ unconditionally. The fact motivates their converting the VAE-based residual coder into a conditional VAE by concatenating $x_c$ and $x_t$ for encoding, and their latent representations for decoding. The idea was extended in~\cite{iclrw} for conditional motion coding, which encodes motion latents in an implicit, one-stage manner. However, Fabian~\emph{et al.}~\cite{FAU} show that these conditional VAE-based approaches~\cite{mmsp,iclrw} may suffer from the \textit{bottleneck} issue; that is, the latent representation of $x_c$ produced by a neural network for conditional decoding may not capture all the information of $x_c$, which serves as a condition for encoding $x_t$. Such information loss and asymmetry can harm the efficiency of conditional coding. Li~\emph{et al.}~\cite{dcvc} improve the work in~\cite{mmsp} by ensuring that the same information-rich latent representation of $x_c$ is utilized for both conditional encoding and decoding. Likewise, the work in~\cite{accv} creates the same coding context for conditional encoding and decoding via a feedback recurrent module that aggregates the past latent information. In common, these approaches do not evaluate any residual signal explicitly.



\begin{figure}[t!]
\begin{center}
    \begin{subfigure}{0.26\linewidth}
        \centering
        \includegraphics[width=\linewidth]{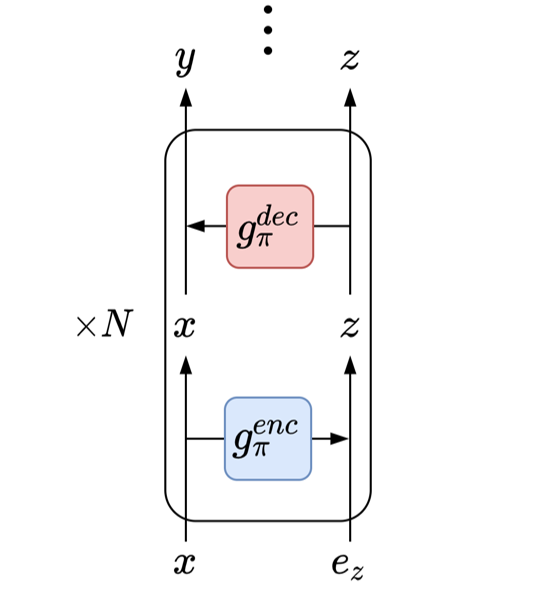}
        \caption{}
        \label{fig:anf_a}
    \end{subfigure}
    \begin{subfigure}{0.26\linewidth}
        \centering
        \includegraphics[width=\linewidth]{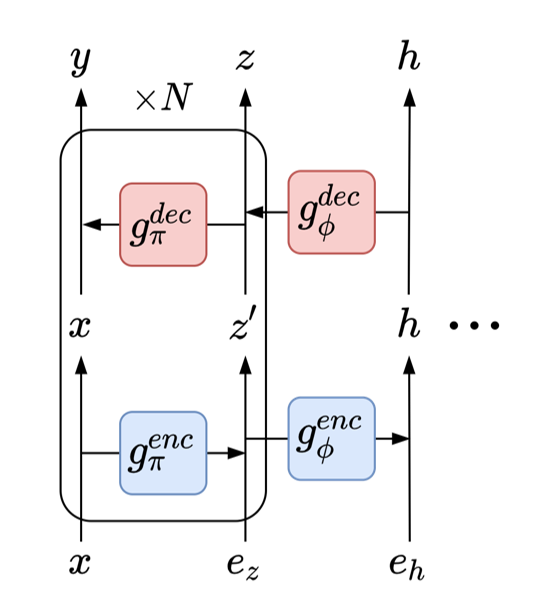}
        \caption{}
        \label{fig:anf_b}
    \end{subfigure}
    \begin{subfigure}{0.26\linewidth}
        \centering
        \includegraphics[width=\linewidth]{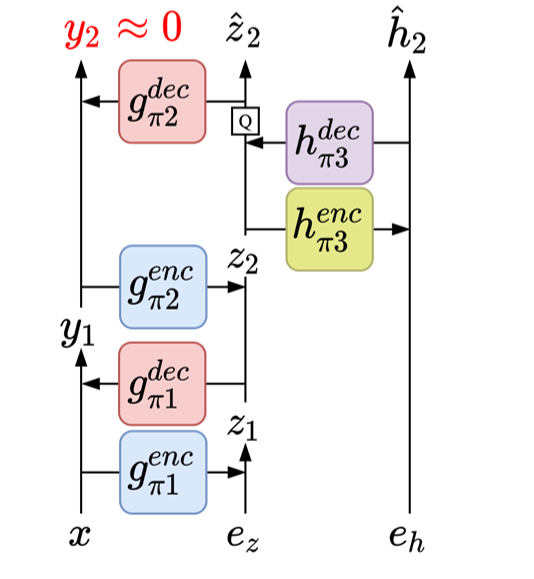}
        \caption{}
        \label{fig:anf_c}
    \end{subfigure}
\vspace{-0.3em}
\caption{The architectures of ANF: (a) N-step ANF, (b) N-step hierarchical ANF, and (c) ANF for image compression (ANFIC).}
\label{fig:anf}
\vspace{-2.5em}
\end{center}
\end{figure}

\subsection{Augmented Normalizing Flows (ANF)}
\label{Sec:ANF}
To learn properly the conditional distribution $p(x_t|x_c)$ for conditional coding, we turn to augmented normalizing flows (ANF), a special type of generative model able to achieve superior expressiveness to VAE. Different from the vanilla flow models~\cite{normalizingflows,glow,realnvp}, ANF~\cite{anf} augments the input $x$ with an independent noise $e_z$ (Fig.~\ref{fig:anf_a}), allowing the augmented noise to induce a complex marginal on $x$~\cite{anf}. ANF contains the autoencoding transform $g_\pi$ as a basic building block, where the encoding transformation $g^{enc}_\pi$ from $(x,e_z)$ to $(x,z)$ and the decoding transformation $g^{dec}_\pi$ from $(x,z)$ to $(y,z)$ are specified by
\begin{align}
    & g^{enc}_\pi(x,e_z)=(x, s^{enc}_\pi(x) \odot e_z + m^{enc}_\pi(x))=(x,z), \label{eq:ANF_1} \\
    & g^{dec}_\pi(x,z)=((x-\mu^{dec}_\pi(z))/\sigma^{dec}_\pi(z), z)=(y,z), \label{eq:ANF_2}
\end{align}
respectively. $s_{\pi}^{enc}$, $m_{\pi}^{enc}$, $\mu_{\pi}^{dec}$ and $\sigma_{\pi}^{dec}$ are element-wise affine transform parameters, and they are driven by neural networks parameterized by $\pi$.

The training of ANF aims to maximize the \textit{augmented data likelihood}--namely, $\arg \max_{\pi} p_{\pi}(x,e_z)=p(g_{\pi}(x,e_z))|det(\partial g_{\pi}(x,e_z)/\partial (x,e_z))|$.
Performing one autoencoding transformation $(y,z)=g_{\pi}(x,e_z)$ (known as the one-step ANF) is equivalent to training a VAE by maximizing the evidence lower bound (ELBO) on the log-marginal $\log p_{\pi}(x)$~\cite{vae}. As such, the learned image compression with the factorized prior~\cite{iclr17balle} can be viewed as an one-step ANF that adopts a purely additive autoencoding transform (i.e.~$s_{\pi}^{enc}(x)=\sigma^{dec}_\pi(z)=1$) and an augmented noise $e_z \sim \mathcal{U}(-0.5,0.5)$ modeling the uniform quantization. In this case, the latents $y,z$ follow the standard Normal $\mathcal{N}(0,I)$ and the learned factorized prior, respectively. In particular, the hyperprior extension~\cite{googleiclr18} has a similar structure to the \textit{hierarchical ANF} (Fig.~\ref{fig:anf_b}), an enhanced form of ANF~\cite{anf} with $g_{\phi}^{enc},g_{\phi}^{dec}$ playing a similar role to the hyper codec. For better expressiveness, one can stack multiple one-step ANF's as the \textit{multi-step ANF}. In~\cite{anfic}, Ho \emph{et al.} introduce the first ANF-based image compression (Fig~\ref{fig:anf_c}), which combines the multi-step and the hierarchical ANF's. 



\vspace{-0.3em}
\section{Proposed Method}
\label{sec:method}

\vspace{-0.3em}
\subsection{Problem Statement}
\vspace{-0.1em}
\label{sec:Statement}
In this section, we formally define our task and objective. Let $x_{1:T} \in \mathbb{R}^{T \times 3 \times H \times W}$ denote a (RGB) video sequence of width $W$ and height $H$ to be encoded, and $\hat{x}_{1:T}$ the decoded video. The video compression task is to strike a good balance between the distortion $d(\hat{x}_{1:T}, x_{1:T})$ of the decoded video $\hat{x}_{1:T}$ and the rate $r(\hat{x}_{1:T})$ needed to represent it. When $T=1$, the task reduces to image compression, of which the problem is cast as learning a VAE by maximizing the ELBO on the log-likelihood $\log p(x)$ in~\cite{googleiclr18}. The same perspective is applicable to video compression yet with the aim of learning a VAE that maximizes the joint log-likelihood $\log p(x_{1:T})$. Because $p(x_{1:T})$ factorizes as $\prod_{t=1}^T p(x_t|x_{<t})$, with $x_{<t}$ representing collectively the video frames up to time instance $t-1$, video compression is often done frame-by-frame by learning the conditional distribution $p(x_t|x_{<t})$. In our task, the decoded frames $\hat{x}_{<t}$ are used in place of $x_{<t}$.

With the traditional predictive coding framework, the ELBO on $\log p(x_t|\hat{x}_{<t})$ has a form of 
\begin{align}
     E_{q(\hat{f}_t,\hat{r}_t|x_t,\hat{x}_{<t})} \log p(x_t|\hat{f}_t,\hat{r}_t,\hat{x}_{<t}) 
     -D_{KL}(q(\hat{f}_t,\hat{r}_t|x_t,\hat{x}_{<t})||p(\hat{f}_t,\hat{r}_t|\hat{x}_{<t})), \label{eq:ELBO_VC}
\end{align}
where the latents $\hat{f}_t \in \mathbb{R}^{2 \times H \times W},\hat{r}_t \in \mathbb{R}^{3 \times H \times W}$ represent the (quantized) optical flow map and the (quantized) motion-compensated residual frame associated with $x_t \in R^{3 \times H \times W}$, respectively. The encoding distribution $q(\hat{f}_t,\hat{r}_t|x_t,\hat{x}_{<t})=q(\hat{f}_t|x_t,\hat{x}_{<t})q(\hat{r}_t|\hat{f}_t,x_t,\hat{x}_{<t})$ specifies the generation of $\hat{f}_t,\hat{r}_t$, while the decoding distribution $p(x_t|\hat{f}_t,\hat{r}_t,\hat{x}_{<t})=\mathcal{N}(\hat{r}_t+warp(\hat{x}_{t-1};\hat{f}_t),\frac{1}{2\lambda} I)$ models the reconstruction process of $x_t$, with $warp(\hat{x}_{t-1}; \hat{f}_t$) denoting the backward warping of $\hat{x}_{t-1}$ based on $\hat{f}_t$, and $\frac{1}{2\lambda}$ being the variance of the Gaussian. Assuming the use of uniform quantization function for obtaining $\hat{f}_t,\hat{r}_t$, the Kullback–Leibler (KL) divergence $D_{KL}(\cdot||\cdot)$ evaluates to the rate costs associated with their transmission: 
\begin{align}
    & D_{KL}(q(\hat{f}_t,\hat{r}_t|x_t,\hat{x}_{<t})||p(\hat{f}_t,\hat{r}_t|\hat{x}_{<t})) \label{eq:ELBO_KL} \\
    & = E_{q(\hat{f}_t,\hat{r}_t|x_t,\hat{x}_{<t})}(-\log p(\hat{f}_t|\hat{x}_{<t})
    -\log p(\hat{r}_t|\hat{f}_t,\hat{x}_{<t})). \nonumber
\end{align} Substituting Eq.~\eqref{eq:ELBO_KL} into Eq.~\eqref{eq:ELBO_VC} and applying the law of total expectation yields  
\begin{align}
    E_{q(\hat{f}_t|x_t,\hat{x}_{<t})} (RD_r(x_t|\hat{f}_t,\hat{x}_{<t}) + \log p (\hat{f}_t|\hat{x}_{<t})),
    \label{eq:motion_coding}
\end{align} where
\begin{align}
     RD_r(x_t|\hat{f}_t,\hat{x}_{<t})= 
     E_{q(\hat{r}_t|\hat{f}_t,x_t,\hat{x}_{<t})} (\log p(x_t|\hat{f}_t,\hat{r}_t,\hat{x}_{<t}) + \log p(\hat{r}_t|\hat{f}_t,\hat{x}_{<t})), 
     \label{eq:residual_coding}
\end{align} which bears the interpretation of the ELBO on $\log p(x_t|\hat{f}_t,\hat{x}_{<t})$, with the latent being the quantized residual frame $\hat{r}_t$. 

From Eqs.~\eqref{eq:motion_coding} and~\eqref{eq:residual_coding}, we see that the traditional predictive coding of a video frame $x_t$ includes (1) encoding the residual frame $\hat{r}_t$ based on $\hat{f}_t,\hat{x}_{<t}$ in order to maximize the log-likelihood $\log p(x_t|\hat{f}_t,\hat{x}_{<t})$ and (2) encoding the flow map $\hat{f}_t$ in a way that strikes a good balance between the maximization of $\log p(x_t|\hat{f}_t,\hat{x}_{<t})$ and the (negative) rate $E_{q(\hat{f}_t|x_t,\hat{x}_{<t})}\log p (\hat{f}_t|\hat{x}_{x<t})$ needed to signal $\hat{f}_t$.  

In this work, we propose to turn the maximization of the log-likelihood $\log p(x_t|\hat{f}_t,\hat{x}_{<t})$, i.e.~Eq.~\eqref{eq:residual_coding}, into a problem of conditional coding, where $\hat{f}_t,\hat{x}_{<t}$ are utilized to formulate the motion-compensated frame $x_c \in \mathbb{R}^{3 \times H \times W}$ as a condition. Unlike the existing works~\cite{mmsp,iclrw,dcvc,accv}, which adopt the conditional VAE, our conditional coder is constructed based on multi-step CANF in modeling $p(x_t|x_c)$ for its better expressiveness.




\vspace{-0.3em}
\subsection{System Overview}


Fig.~\ref{fig:network} depicts our CANF-based video compression system, abbreviated as CANF-VC. It includes two major components: (1) the CANF-based inter-frame coder $\{G_{\pi},G_{\pi}^{-1}\}$ and (2) the CANF-based motion coder $\{F_{\pi},F_{\pi}^{-1}\}$. The inter-frame coder encodes a video frame $x_t$ conditionally, given the motion-compensated frame $x_c$. It departs from the conventional residual coding by maximizing the conditional log-likelihood $p(x_t|x_c)$ with CANF model (Section~\ref{sec:inter_cond}). The motion coder shares a similar architecture to the inter-frame coder. It extends conditional coding to motion coding, in order to signal the flow map $f_t$, which characterizes the motion between $x_t$ and its reference frame $\hat{x}_{t-1}$. In our work, $f_t$ is estimated by PWC-Net~\cite{pwc}. The compressed flow map $\hat{f}_t$ serves to warp the reference frame $\hat{x}_{t-1}$, with the warped result enhanced further by a motion compensation network to arrive at $x_c$. To formulate a condition for conditional motion coding, we introduce a flow extrapolation network to extrapolate a flow map $f_c$ from three previously decoded frames $\hat{x}_{t-1}, \hat{x}_{t-2}, \hat{x}_{t-3}$ and two decoded flow maps $\hat{f}_{t-1}, \hat{f}_{t-2}$. Note that we expand the condition of $p(x_t|\hat{x}_{<t})$ from previously decoded frames $\{\hat{x}_{<t}\}$ to include also previously decoded flows $\{\hat{f}_{<t}\}$.

\vspace{-0.3em}
\subsection{CANF-based Inter-frame Coder}
\label{sec:inter_cond}
\vspace{-0.3em}

\begin{figure}[t!]
\begin{center}
    \begin{subfigure}{0.55\linewidth}
        \centering
        \includegraphics[width=\textwidth]{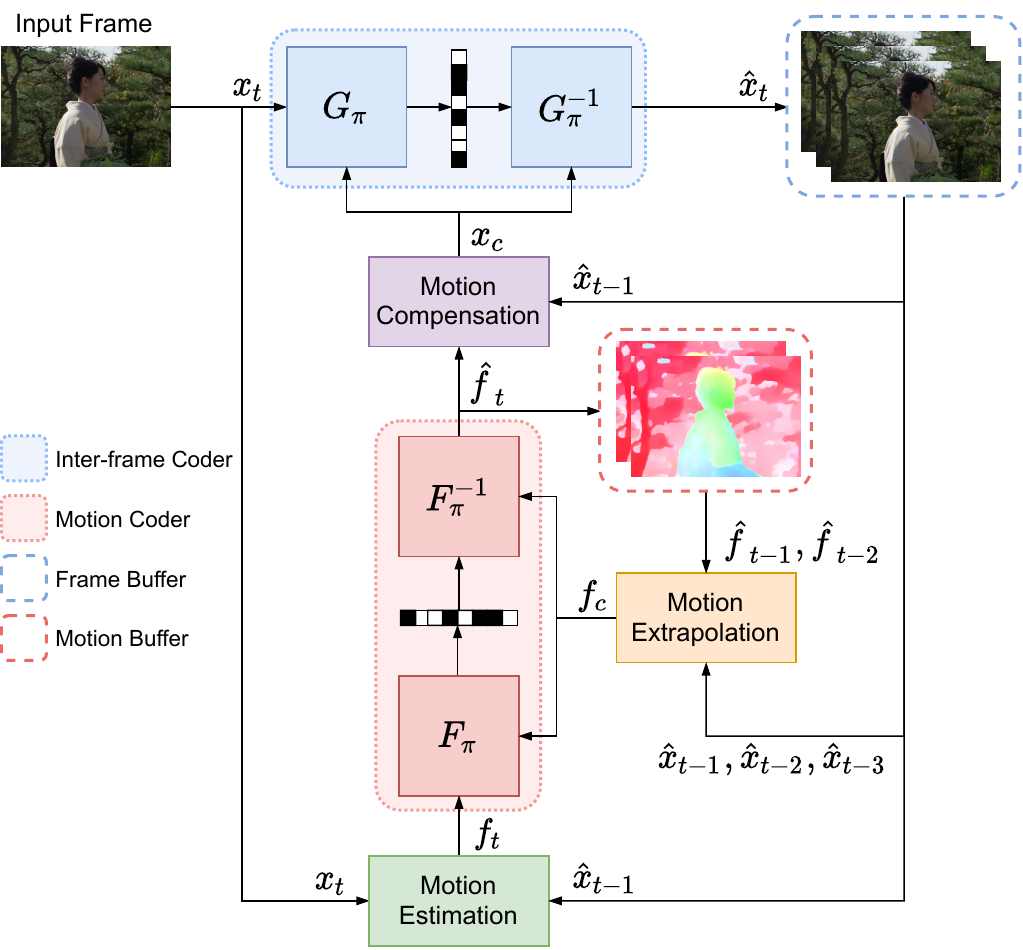}
        \caption{}
        \label{fig:network}
    \end{subfigure}
    \begin{subfigure}{0.43\linewidth}
        \vspace{1.5em}
        \centering
        \includegraphics[width=0.9\textwidth]{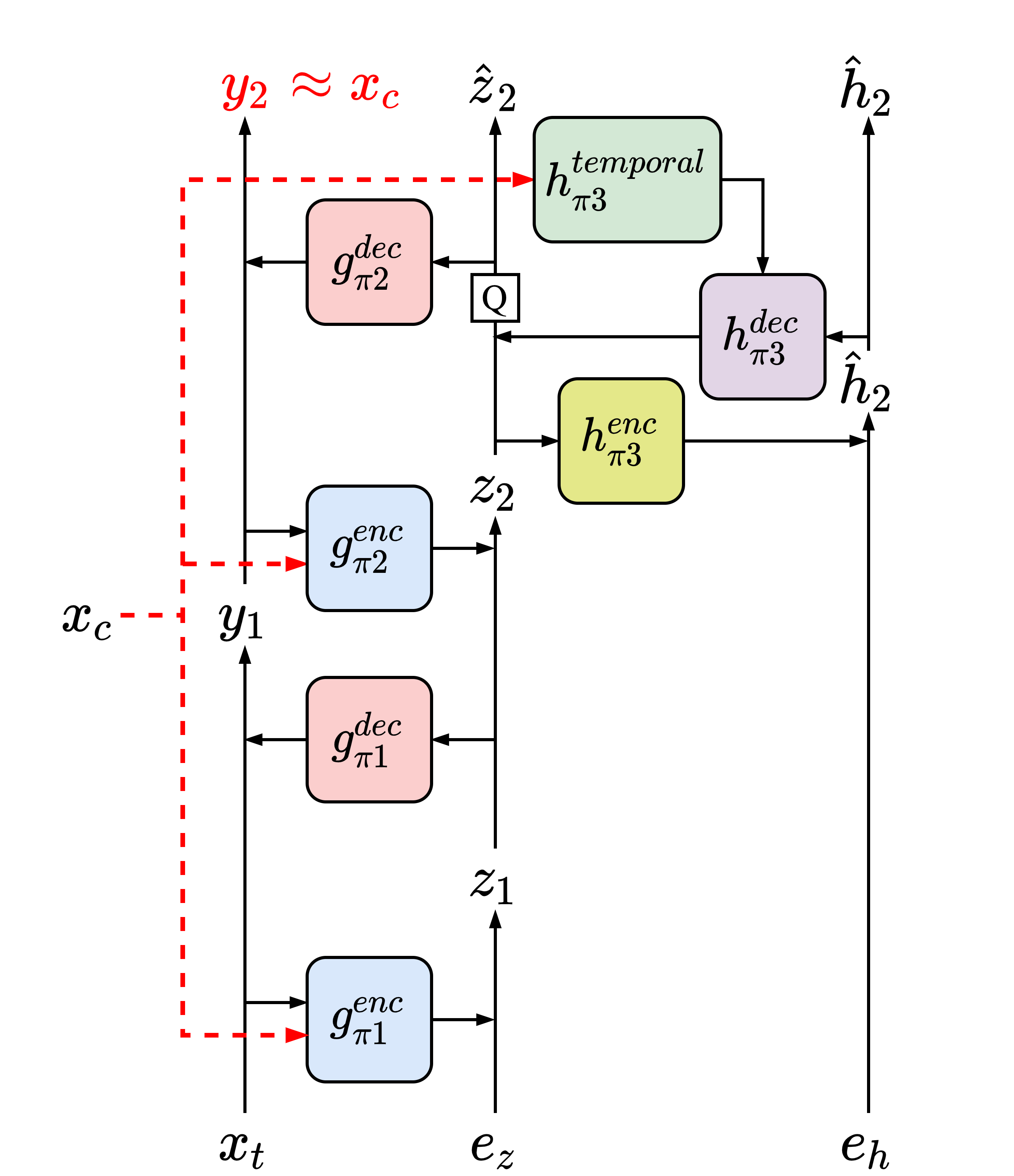}
        \vspace{0.63em}
        \caption{}
        \label{fig:canf}
    \end{subfigure}

\caption{Illustration of (a) the proposed CANF-VC framework and (b) the CANF-based inter-frame coder $\{G_\pi,G_{\pi}^{-1}\}$. The CANF-based motion coder $\{F_\pi,F_{\pi}^{-1}\}$ follows the same design as the inter-frame coder, with $x_t, x_c$ replaced by $f_t, f_c$, respectively.}
\vspace{-3em}
\end{center}
\end{figure}

Fig.~\ref{fig:canf} presents the architecture of our CANF-based inter-frame coder, which aims to learn the conditional distribution $p(x_t|x_c)$ of the coding frame $x_t$ given the motion-compensated frame $x_c$. This is achieved by maximizing the augmented likelihood $p(x_t,e_z,e_h|x_c)$ in the CANF framework, where $e_z \in \mathbb{R}^{C \times \frac{H}{16} \times \frac{W}{16}}$, $e_h \in \mathbb{R}^{C \times  \frac{H}{64} \times \frac{W}{64}}$ are the two augmented noise inputs. It is shown in~\cite{anf} that maximizing $p(x_t,e_z,e_h|x_c)$ is equivalent to maximizing a lower bound on the marginal likelihood $p(x_t|x_c)$. 


\textbf{Architecture:} Motivated by~\cite{anfic}, our conditional inter-frame coder is a hybrid of the two-step and the hierarchical ANF's. The two autoencoding transforms $\{g_{\pi_1}^{enc},g_{\pi_1}^{dec}\},\{g_{\pi_2}^{enc},g_{\pi_2}^{dec}\}$ convert $x_t,e_z$ into their latents $y_2,z_2$, respectively, while the hierarchical autoencoding transform $\{h_{\pi_3}^{enc},h_{\pi_3}^{dec}\}$ acts as the hyperprior codec, encoding the latent $z_2$ into the hyperprior representation $\hat{h}_2$. The volume preserving property of CANF requires that the latents $y_2,z_2~(or~\hat{z}_2),\hat{h}_2$ have the same dimensions as their respective inputs $x_t,e_z,e_h$. One notable distinction between CANF and ANFIC~\cite{anfic} is the incorporation of the condition $x_c$ into the autoencoding transforms and the prior distribution, as will be detailed next.


\textbf{Conditional Encoding:} The core idea of our conditional coding is to let the latent $y_2$, which represents a transformed version of the target frame $x_t$, approximate the condition $x_c$, with the latents $z_2,\hat{h}_2$ encoding the information necessary for instructing the transformation. Specifically, the two autoencoding transforms operate similarly and successively. Taking $\{g_{\pi_1}^{enc},g_{\pi_1}^{dec}\}$ as an example (see Fig.\ref{fig:canf}), we have 
\begin{align}
    & g^{enc}_{\pi_1}(x_t,e_z|x_c)=(x_t, e_z + m^{enc}_{\pi_1}(x_t, x_c))=(x_t,z_1), 
    \\
    & g^{dec}_{\pi_1}(x_t,z_1)=(x_t-\mu^{dec}_{\pi_1}(z_1), z_1)=(y_1,z_1). 
\end{align}
That is, the one-step transformation from $x_t$ to $y_1$ is done by subtracting the decoder output $\mu^{dec}_{\pi_1}(z_1)$ from $x_t$. Note that $\mu^{dec}_{\pi_1}(z_1)$ decodes the latent $z_1$, which aggregates the information of $x_t,x_c$, and the augmented noise $e_z$. We remark that the encoding process is made conditional on $x_c$ by concatenating $x_c$ and $x_t$ to form the encoder input. Intuitively, supplying 
$x_c$ as an auxiliary signal should ease the transformation from $x_t$ to $x_c$. This process is repeated by taking $y_1$ and $z_1$ as the inputs to the next autoencoding transform $\{g_{\pi_2}^{enc},g_{\pi_2}^{dec}\}$. In fact, the number of the autoencoding transforms is flexible. In comparison with Eqs.~\eqref{eq:ANF_1} and~\eqref{eq:ANF_2}, our autoencoding transform is purely additive (i.e. $s_{\pi}^{enc},\sigma_{\pi}^{dec}$ in Eqs.~\eqref{eq:ANF_1} and~\eqref{eq:ANF_2} are set to 1), which is found beneficial in terms of training stability.

The hierarchical autoencoding transform $\{h_{\pi_3}^{enc},h_{\pi_3}^{dec}\}$ serves to estimate the probability distribution of $z_2$ for entropy coding. It operates according to 
\begin{align}
    & h^{enc}_{\pi_3}(z_2,e_h)=(z_2, e_h + m^{enc}_{\pi_3}(z_2))=(z_2,\hat{h}_2), 
    \\
    & h^{dec}_{\pi_3}(z_2,\hat{h}_2|x_c)=(\lfloor z_2 - \mu_{\pi_3}^{dec}(\hat{h}_2, h_{\pi_3}^{temporal}(x_c))  \rceil, \hat{h}_2)=(\hat{z}_2, \hat{h}_2), 
\end{align}
where $\lfloor \cdot \rceil$ (depicted as Q in Fig.~\ref{fig:canf}) denotes the nearest-integer rounding, which is needed to express $z_2$ in fixed-point representation for lossy compression. At training time, the rounding effect is modeled by additive quantization noise. It is worth noting that $x_c$ is provided as an auxiliary input to $\mu_{\pi_3}^{dec}$ to exert a combined effect of the hyperprior and the temporal prior ($h_{\pi_3}^{temporal}$ in Fig.~\ref{fig:canf}).

\textbf{Conditional Decoding:} The decoding process of our inter-frame coder updates the motion-compensated frame $x_c$ successively to reconstruct $x_t$. It starts by entropy decoding the latents $\hat{z}_2,\hat{h}_2$, and substituting $x_c$ for $y_2$. The quantized $z_2$ will then be recovered and decoded to reconstruct $\mu^{dec}_{\pi_2}(z_2)$, which updates $x_c$ as $y_1=x_c+\mu^{dec}_{\pi_2}(z_2)$. Subsequently, $y_1$ will be encoded conditionally based on $x_c$ using $m^{enc}_{\pi_2}(y_1,x_c)$ in order to update the latent $z_2$ as $z_1=z_2-m^{enc}_{\pi_2}(y_1,x_c)$. Finally, $z_1$ is decoded by $\mu^{dec}_{\pi_1}(z_1)$ to update $y_1$ as the reconstructed version $\hat{x}_t=y_1+\mu^{dec}_{\pi_1}(z_1)$ of $x_t$. In a sense, the reconstruction of $x_t$ is achieved by passing the latent $\hat{z}_2$ through the composition of the decoding and encoding transforms to update $x_c$.

\textbf{Conditional Prior Distribution:} Another strategy we adopt to learn $p(x_t,e_z,e_h|x_c)$ is to introduce a conditional prior distribution $p(y_2, \hat{z}_2, \hat{h}_2 | x_c)$. Specifically, we assume that it factorizes as follows:
\begin{equation}
    p(y_2, \hat{z}_2, \hat{h}_2 | x_c) = p(y_2| x_c)p(\hat{z}_2 | \hat{h}_2, x_c) p(\hat{h}_2).
\label{eq:entropy_inter}
\end{equation}
Because we require $y_2$ to approximate $x_c$, it is natural to choose $p(y_2|x_c)$ to be $\mathcal{N}(x_c, \frac{1}{2\lambda_1} I)$ with a small variance $\frac{1}{2\lambda_1}$. Moreover, following the hyperprior~\cite{googleiclr18}, $p(\hat{z}_2|\hat{h}_2, x_c)$ and $p(\hat{h}_2)$ are modeled by
\begin{equation}
\begin{split}
     p(\hat{z}_2|\hat{h}_2, x_c) &= \mathcal{N}(0,(\sigma^{dec}_{\pi_3}(\hat{h}_2, h_{\pi_3}^{temporal}(x_c)))^2 I) \ast \mathcal{U}(-0.5,0.5) \\
     p(\hat{h}_2) &= \mathcal P_{\hat{h}_2|\psi} \ast \mathcal{U}(-0.5,0.5), \nonumber
\end{split}
\end{equation}
where $\ast$ denotes convolution and $P_{\hat{h}_2|\psi}$ is a factorized prior parameterized by $\psi$. The use of the motion-compensated frame $x_c$ along with $\hat{h}_2$ in estimating the distribution of $\hat{z}_2$ combines temporal prior ($h_{\pi_3}^{temporal}$ in Fig.~\ref{fig:canf}) and hyperprior. 

\textbf{Augmented Noises $e_z,e_h$:} In the theory of ANF~\cite{anf}, the augmented noises are meant to induce a complex marginal on the input $x$. For the compression task, we fix $e_z$ at 0 during training and test, in order not to increase the entropy rate at $\hat{z}_2$. For training, the quantization Q in Fig.~\ref{fig:canf} is simulated by additive noise. In contrast, we draw $e_h \sim \mathcal{U}(-0.5,0.5)$ for simulating the quantization of the hyperprior at training time, and set it to zero at test time when the hyperprior is actually rounded. 



\textbf{Extension to Conditional Motion Coding:} The CANF-based motion coder follows the same design as the CANF-based inter-frame coder. The coding frame $x_t$ is replaced with the optical flow map $f_t$ and the motion-compensated frame $x_c$ with the extrapolated flow map $f_c$. In addition, the temporal prior takes the extrapolated frame $warp(\hat{x}_{t-1}; f_c)$ as input. To perform the flow map extrapolation, we adopt a U-Net-based network (see supplementary document).

In the supplementary document, we provide another CANF implementation, which additionally accepts $x_c$ as input to the decoding transforms. We choose the current implementation due to its comparable performance and simpler design.

\vspace{-0.3em}
\subsection{Training Objective}

We train the conditional inter-frame and motion coders end-to-end. Inspired by Eq.~\eqref{eq:motion_coding}, we first turn the maximization of the ELBO (i.e.~$RD_r$ in Eq.~\eqref{eq:residual_coding}) on $\log  p(x_t|\hat{f}_t,\hat{x}_{<t})$ into maximizing $\log p(x_t,e_z,e_h|x_c)$. That is, to minimize
\begin{equation}
\begin{split}
    -\log p(x_t,e_z,e_h|x_c) & = 
    -\log p(\hat{h}_2) -\log p(\hat{z}_2|\hat{h}_2, x_c) \\ + \lambda_1 \|y_2-x_c\|^2 
    & -log\left|det\frac{\partial G_{\pi }(x_t,e_z,e_h|x_c)}{\partial (x_t,e_z,e_h)}\right|.   \nonumber
\label{eq:weighted_prediction}
\end{split}
\end{equation} 
To ensure the reconstruction quality, we follow~\cite{anfic} to replace the negative log-determinant of the Jacobian with a weighted reconstruction loss $\lambda_2 d(x_t,\hat{x}_t)$, arriving at
\begin{equation}
\begin{split}
    -\log p(x_t,e_z,e_h|x_c) & \approx  
    \underbrace{-\log p(\hat{h}_2) -\log p(\hat{z}_2|\hat{h}_2, x_c)}_R 
     + \lambda_1 \|y_2-x_c\|^2 
     + \underbrace{\lambda_2 d(x_t,\hat{x}_t)}_D,  \nonumber
\label{eq:training_objective}
\end{split}
\vspace{-1em}
\end{equation} 
which includes the rate $R$ needed to signal the transformation between $x_t$ and $x_c$, the regularization term requiring $y_2$ to approximate $x_c$, and the distortion $D$ of $\hat{x}_t$. To complete the loss function, we also follow the second term in Eq.~\eqref{eq:motion_coding} to include the conditional motion rate used to signal $f_t$ given $f_c$, which leads to
\begin{equation}
\begin{split}
      \mathcal{L} = 
     & -\log p(\hat{h}_2) -\log p(\hat{z}_2|\hat{h}_2, x_c) + \lambda_1 \|y_2-x_c\|^2\\
     & -\log p_f(\hat{h}_2) - \log p_f(\hat{z}_2|\hat{h}_2, f_c) 
      + \lambda_2 d(x_t, \hat{x}_t),
\label{eq:oveall_training_objective}
\end{split}
\end{equation} 
where $p_f$ (respectively, $p$) describes the prior distribution over the motion (respectively, inter-frame) latents.

\vspace{-0.3em}
\subsection{Comparison with ANFIC and Other VAE-based Schemes}
\vspace{-0.3em}
Our CANF-VC is based on ANF, as well as ANFIC, a learned image compression system proposed in~\cite{anfic}. However, they significantly differ from each other, not only because they refer to different applications. ANFIC~\cite{anfic} adopts an \textit{unconditional} ANF to learn the image distribution $p(x)$ for image compression, whereas CANF-VC uses two \textit{conditional} ANF's (CANF's) to learn the conditional distribution $p(x_t|x_c)$ for inter-frame coding and the conditional rate needed to signal the motion part, respectively. As a result, CANF-VC is a complete video coding framework. Note that how the conditional information $x_c$ and $f_c$ are both incorporated in the respective autoencoding transforms and in the respective prior distributions is first proposed in this work.

CANF-VC is also distinct from conditional VAE-based frameworks, such as DCVC~\cite{dcvc} and~\cite{iclrw}. CANF-VC bases the compression backbone on CANF, which is a flow-based model and includes VAE as a special case. As compared with DCVC~\cite{dcvc}, CANF-VC additionally features conditional motion coding. 
Although conditional motion coding also appears in~\cite{iclrw}, their VAE-based approach does not explicitly estimate a flow map prior to conditional coding, and may suffer from the bottleneck issue~\cite{FAU} (Section~\ref{subsec:CondCoding}). In contrast, CANF-VC takes an explicit approach and avoids the bottleneck issue by using the same $x_c$ symmetrically in the encoder and the decoder due to its invertible property. 

\vspace{-0.5em}
\section{Experiments}
\vspace{-0.5em}
\label{sec:experiment}


\subsection{Settings and Implementation Details}
\vspace{-0.3em}
\label{subsec:Settings}

\textbf{Training Details:}
We train our model on Vimeo-90k~\cite{vimeo} dataset, which contains 91,701 7-frame sequences with resolution $448 \times 256$. We randomly crop these video clips into $256 \times 256$ for training. We adopt the Adam \cite{Adam} optimizer with the learning rate $10^{-4}$ and the batch size 32. Separate models are trained to optimize first the mean-square error with $\lambda_2=\{256, 512, 1024, 2048\}$ and $\lambda_1 = 0.01*\lambda_2$ (see Eq.~\eqref{eq:oveall_training_objective}). We then fine-tune these models for Multi-scale Structural Similarity Index (MS-SSIM), with $\lambda_2$ set to $\{4, 8, 16, 32, 64\}$. All the low-rate models are adapted from the one trained for the highest rate point.

\textbf{Evaluation Methodologies:} We evaluate our models on commonly used datasets, including UVG~\cite{uvg}, MCL-JCV~\cite{mcl}, and HEVC Class B~\cite{hevcctc}.
We follow common test protocols to provide results in Table~\ref{tab:BDrate_gop12} for 100-frame encoding with GOP \footnote[1]{GOP refers to Group-of-Pictures and is often used interchangeably with the intra period in papers on learned video codecs.} size 10 on HEVC Class B, and full-sequence encoding with GOP size 12 on the other datasets. Additionally, we present results for GOP size 32 in Table~\ref{tab:BDrate_gop32}, to underline the contributions of our inter-frame and motion coders. For this additional setting, all the learned codecs use ANFIC~\cite{anfic} as the intra-frame coder and encode only the first 96 frames in every test sequence. To evaluate the rate-distortion performance, the bit rates are measured in bits per pixel (bpp), and the quality in PSNR-RGB and MS-SSIM-RGB. Moreover, we use x265 in veryslow mode as the anchor for reporting BD-rates. 

\textbf{Baseline Methods:} The baseline methods for comparison include x265, HEVC Test Model (HM)~\cite{HM} and several recent publications, including DVC\_Pro \cite{dvcpro}, M-LVC~\cite{mlvc}, RaFC~\cite{rafc}, FVC~\cite{fvc} and DCVC~\cite{dcvc}. Because these baseline methods adopt different intra-frame coders (see the second column of Table~\ref{tab:BDrate_gop12}), which are critical to the overall rate-distortion performance, we provide results with ANFIC~\cite{anfic} (CANF-VC) and BPG (CANF-VC*) as the intra-frame coders to ease comparison. Note that ANFIC~\cite{anfic} shows comparable performance to cheng2020-anchor~\cite{compressai}. It is to be noted that x265, HM~\cite{HM}, DVC\_Pro \cite{dvcpro}, and M-LVC~\cite{mlvc} use the same model optimized for PSNR to report PSNR-RGB and MS-SSIM-RGB results. While the other methods train separate models in reporting these results. We also present CANF-VC$^-$ and CANF-VC Lite as two additional variants of CANF-VC. CANF-VC$^-$ disables conditional motion coding while CANF-VC Lite implements a lightweight version of CANF-VC by reducing the channels in the autoencoding and the hyperprior transforms, and adopting SPyNet~\cite{spy_net} as the flow estimation network.

\begin{table*}[t]
    \begin{subtable}[ht]{\textwidth}   
    \centering
    \resizebox{\textwidth}{!}{
        \begin{tabular}{lc@{}p{1em}@{}ccc@{}p{1em}@{}ccc@{}p{1em}@{}c}
            \toprule
                & Intra coder   && \multicolumn{3}{c}{BD-rate (\%) PSNR-RGB} && \multicolumn{3}{c}{BD-rate (\%) MS-SSIM-RGB} && \multirow{2}{*}{Size}\\
                    \cline{4-6} \cline{8-10}
                                                     &(PSNR-RGB/MS-SSIM-RGB)       && UVG                              & MCL-JCV                                        & HEVC-B  
                                                                                   && UVG                              & MCL-JCV                          & HEVC-B\\ 
            
            \hline
            DVC\_Pro~\cite{dvcpro}                   & -/-                         &&  -3.0                            &  -                               & -13.1 
                                                                                   && -5.2                             &  -                               & -20.8                           && 29M\\
            M-LVC~\cite{mlvc}                        & BPG/BPG                     && -15.3                            &  18.8                            & -38.6 
                                                                                   && -0.2                             &  4.7                             & -37.9                           && -\\
            RaFC~\cite{rafc}                         & hyperprior/hyperprior       && -11.1                            &  4.4                             & -9.3  
                                                                                   && -25.5                            &  -27.9                           & -37.2                           && -\\
            FVC~\cite{fvc}                           & BPG/BPG                     && -16.9                            &  -3.8                            & -17.8 
                                                                                   && -45.0                            &  -46.1                           & -54.3                           && 26M\\
            HM (LDP, 4 refs)              & -/-                         &&  -29.4                            &  -13.9                           &  -29.6 
                                                                                   && -18.9                &  -13.8                & -17.1                          && -\\  
            \textbf{CANF-VC$^*$}                     & BPG/BPG                     && -35.5                            &  \textcolor{blue}{\textbf{-14.6 }}                          & -35.4 
                                                                                   && -46.6                            & \textcolor{blue}{\textbf{-46.7}} & -53.2                           && 31M\\\hline
            DCVC~\cite{dcvc}                         & cheng2020-anchor/hyperprior && -23.8                            &  -14.4                           & -34.9 
                                                                                   && -43.9                            &  -44.9                           & -50.7                           && 8M \\
            DCVC (ANFIC)                                     & ANFIC/ANFIC                 && -24.8                            &  -13.6                           & -34.0 
                                                                                   && -41.9                            &  -43.7                           & -51.1                           && 8M \\ 
            \textbf{CANF-VC Lite}                    & ANFIC/ANFIC                 && \textcolor{blue}{\textbf{-37.3}} &  -14.3                           & \textcolor{blue}{\textbf{-39.8}} 
                                                                                   && \textcolor{blue}{\textbf{-47.6}} &  -44.2                           & \textcolor{red}{\textbf{-56.8}} && 15M\\
            \textbf{CANF-VC}                         & ANFIC/ANFIC                 && \textcolor{red}{\textbf{-42.5}}  & \textcolor{red}{\textbf{-21.0}}  & \textcolor{red}{\textbf{-40.1}} 
                                                                                   && \textcolor{red}{\textbf{-51.4}}  & \textcolor{red}{\textbf{-47.6}}  & \textcolor{blue}{\textbf{-54.7}}&& 31M\\
            \bottomrule
        \end{tabular}
    }
    \end{subtable}
    \caption {BD-rate comparison with GOP size 10/12. The anchor is x265 in veryslow mode. The best performer is marked in \textcolor{red}{\textbf{red}} and the second best in \textcolor{blue}{\textbf{blue}}.}
    \label{tab:BDrate_gop12} 
    \vspace{-2.5em}  
\end{table*}

\begin{table}[t]
    \begin{subtable}[ht]{\textwidth}
    \centering
    \resizebox{\textwidth}{!}{
        \begin{tabular}{lc@{}p{1em}ccc@{}p{1em}@{}ccc}
            \toprule
                & Intra coder          && \multicolumn{3}{c}{BD-rate (\%) PSNR-RGB} && \multicolumn{3}{c}{BD-rate (\%) MS-SSIM-RGB}\\
                    \cline{3-6} \cline{8-10}
                &(PSNR-RGB/MS-SSIM-RGB) &&  UVG   & MCL-JCV & HEVC-B && UVG & MCL-JCV & HEVC-B\\\hline
            M-LVC (ANFIC)                   & ANFIC/ANFIC &&  -12.1&  -5.3   &  -9.7  &&  -7.5 &  -8.4   &  -18.8\\
            DCVC (ANFIC)                    & ANFIC/ANFIC &&  -16.3&  -21.3  &  -10.5 &&  \textcolor{blue}{\textbf{-38.8}}&  \textcolor{blue}{\textbf{-48.9}}  &  -39.3 \\\hline
            \textbf{CANF-VC Lite}           & ANFIC/ANFIC && \textcolor{blue}{\textbf{-36.1}}&  -26.5  &  \textcolor{blue}{\textbf{-30.3}} &&  -37.9&  -47.8  &  \textcolor{red}{\textbf{-44.2}} \\
            \textbf{CANF-VC$^-$}            & ANFIC/ANFIC &&  -31.1&  -29.5  &  -23.6 &&  -36.2&  -47.8  &  -38.0 \\
            \textbf{CANF-VC}                & ANFIC/ANFIC &&  -35.9&  \textcolor{blue}{\textbf{-32.0}}  &  -27.7 &&  \textcolor{red}{\textbf{-40.3}}&  \textcolor{red}{\textbf{-49.6}}  &  \textcolor{blue}{\textbf{-41.3}} \\\hline
            HM (LDP, 4 refs)        & -/- &&  \textcolor{red}{\textbf{-41.6}}&  \textcolor{red}{\textbf{-38.6}}  &  \textcolor{red}{\textbf{-32.1}} &&  -34.3&  -32.0  &  -31.0 \\ 
            \bottomrule
        \end{tabular}
        
    }
    
    \end{subtable}
    \caption {BD-rate comparison with GOP size 32. All the competing methods (except HM) use ANFIC~\cite{anfic} as the intra-frame coder. The anchor is x265 in veryslow mode. The best performer is marked in \textcolor{red}{\textbf{red}} and the second best in \textcolor{blue}{\textbf{blue}}.}
    \label{tab:BDrate_gop32}
\end{table}

\begin{figure}[t!]
\vspace{-0.3em}
\begin{center}
\begin{subfigure}{0.45\linewidth}
    \centering
    \includegraphics[width=\linewidth]{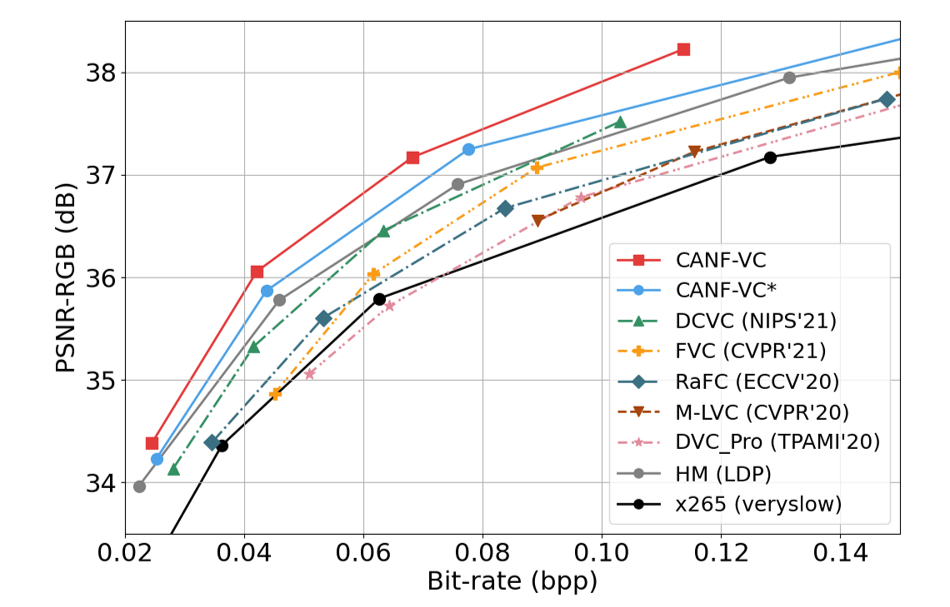} 
    \vspace{-1.5em}
    \caption{UVG, PSNR-RGB}
    \label{fig:uvgPSNR}
\end{subfigure}
\begin{subfigure}{0.45\linewidth}
    \centering
    \includegraphics[width=\linewidth]{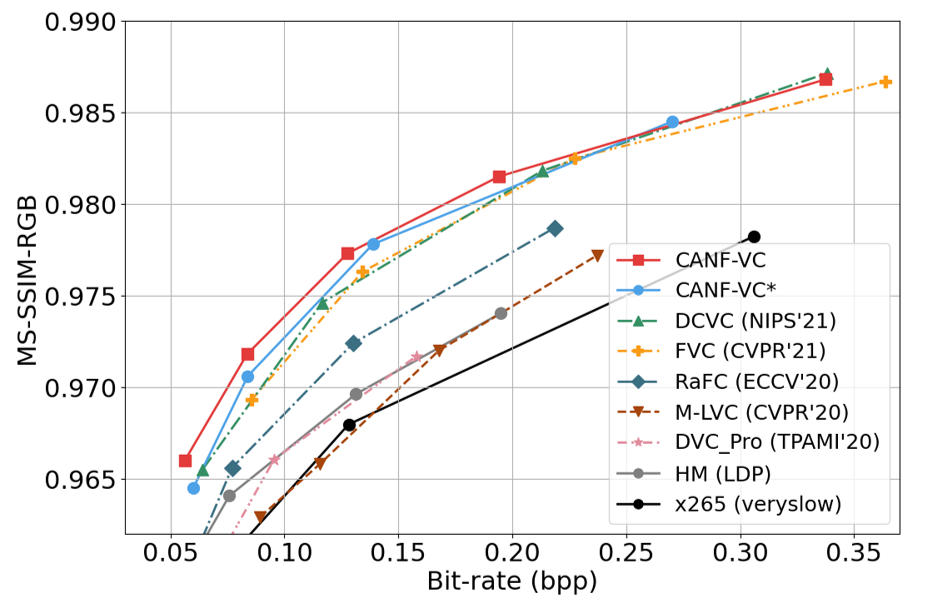}
    \vspace{-1.5em}
    \caption{UVG, MS-SSIM-RGB}
    \label{fig:uvgSSIM}
\end{subfigure}
\begin{subfigure}{0.45\linewidth}
    \centering
    \includegraphics[width=\linewidth]{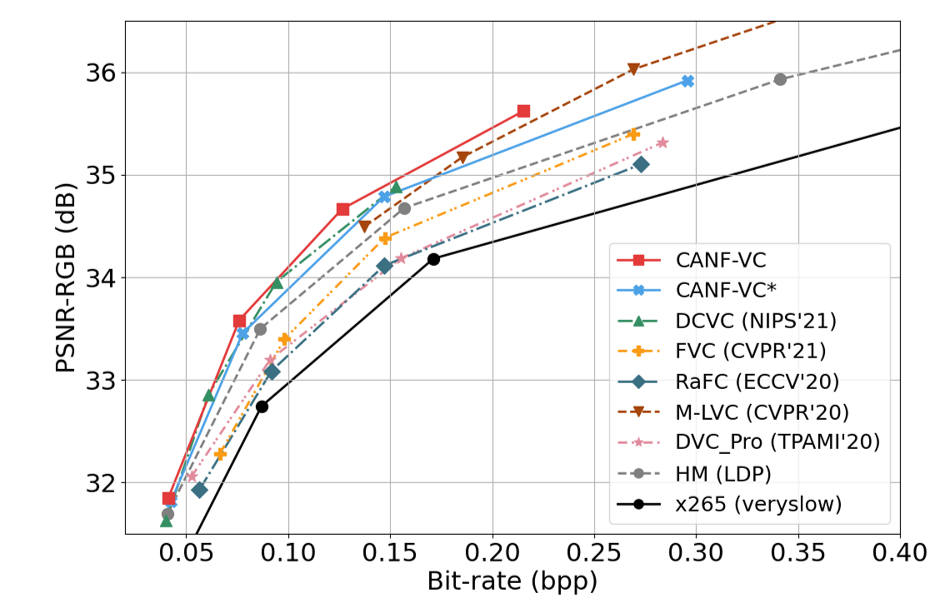}
    \vspace{-1.5em}
    \caption{HEVC Class B, PSNR-RGB}
    \label{fig:hevcPSNR}
\end{subfigure}
\begin{subfigure}{0.45\linewidth}
    \centering
    \includegraphics[width=\linewidth]{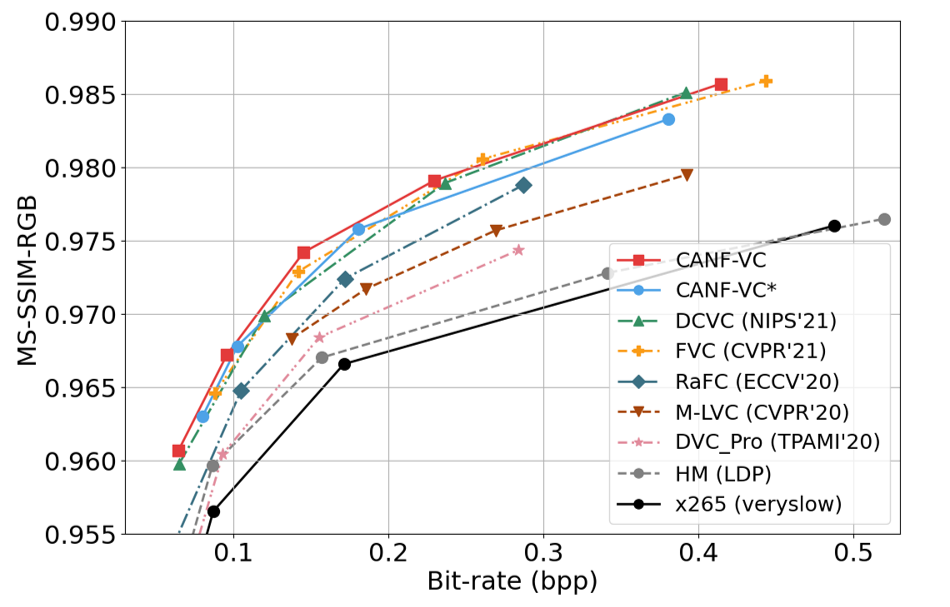}
    \vspace{-1.5em}
    \caption{HEVC Class B, MS-SSIM-RGB}
    \label{fig:hevcSSIM}
\end{subfigure}
\begin{subfigure}{0.45\linewidth}
    \centering
    \includegraphics[width=\linewidth]{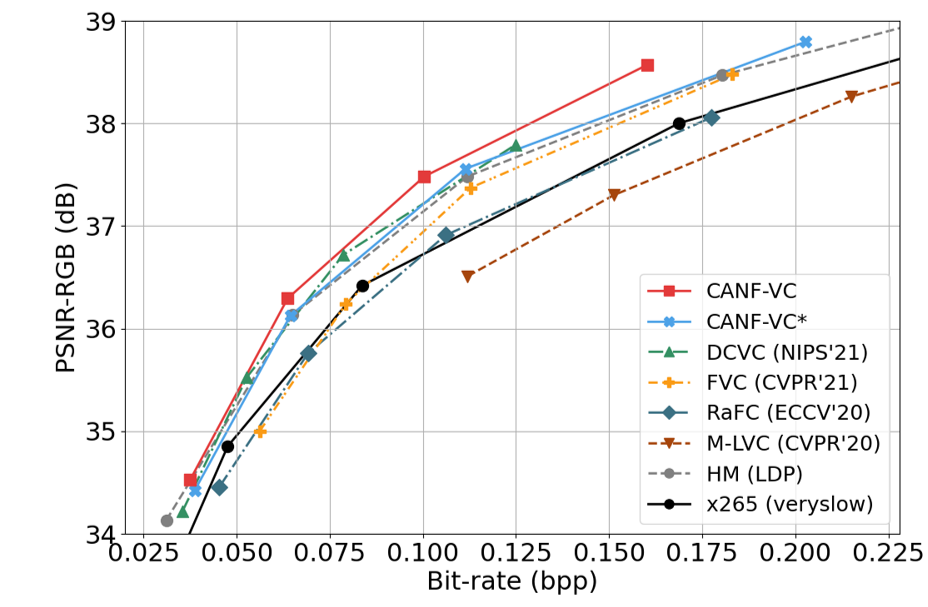}
    \vspace{-1.5em}
    \caption{MCL-JCV, PSNR-RGB}
    \label{fig:mclPSNR}
\end{subfigure}
\begin{subfigure}{0.45\linewidth}
    \centering
    \includegraphics[width=\linewidth]{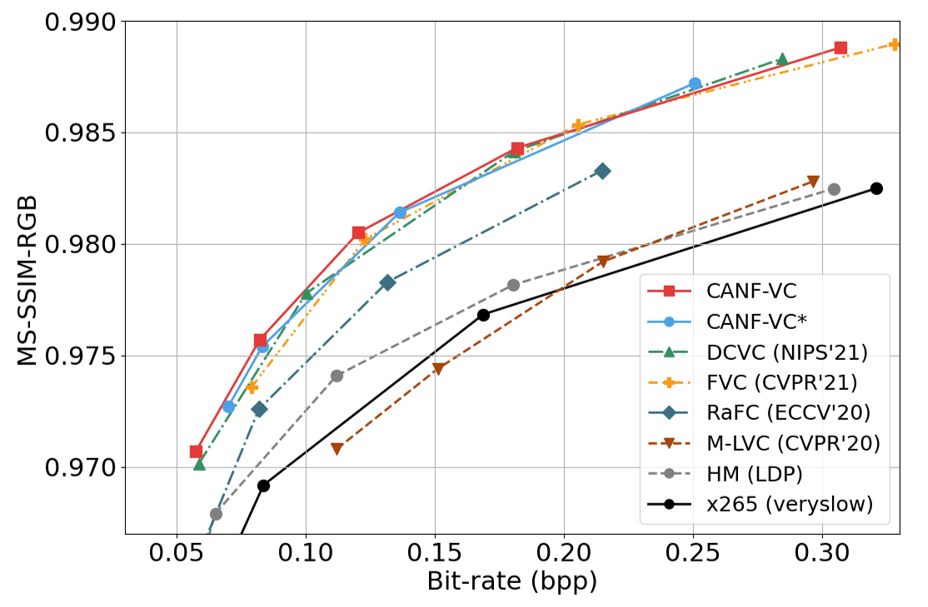}
    \vspace{-1.5em}
    \caption{MCL-JCV, MS-SSIM-RGB}
    \label{fig:mclSSIM}
\end{subfigure}

\vspace{-0.5em}
\caption{Rate-distortion performance evaluation with GOP size 10/12 on UVG, HEVC Class B, and MCL-JCV datasets for both PSNR-RGB and MS-SSIM-RGB.} 
\label{fig:RD}
\vspace{-1.0em}
\end{center}
\vspace{-1.0em}
\end{figure}

\vspace{-0.3em}
\subsection{Rate-Distortion and Subjective Quality Comparison}
\label{subsec:rd}
\vspace{-0.3em}

\textbf{Rate-Distortion Comparison:} The upper part of Table~\ref{tab:BDrate_gop12} compares the competing methods with their intra-frame coders, e.g. hyperprior~\cite{iclr17balle}, performing comparably to BPG. We see that our CANF-VC* (with BPG as the intra-frame coder) outperforms most of these baselines across different datasets in terms of PSNR-RGB. Its slight rate inflation (3\%) as compared to M-LVC~\cite{mlvc} on HEVC-B class may be attributed to the not-fully-aligned rate range in which the BD-rate is measured (see Fig.~\ref{fig:RD}). Note that M-LVC~\cite{mlvc} is initially trained for GOP size 100. With no access to its training software, a rate shift occurs when its test code is re-run for GOP size 10/12. Another observation is that CANF-VC* shows similar MS-SSIM-RGB results to FVC~\cite{fvc}, while surpassing the others considerably. The lower part of Table~\ref{tab:BDrate_gop12} further shows that in terms of both quality metrics, our full model CANF-VC performs consistently better than both DCVC variants, where one uses ANFIC~\cite{anfic} and the other adopts cheng2020-anchor~\cite{compressai} as their respective intra-frame coders. The same observation can be made with CANF-VC$^-$ and CANF-VC Lite, except that they perform similarly to DCVC~\cite{dcvc} on MCL-JCV.

Under the long GOP setting (Table~\ref{tab:BDrate_gop32}), the gain of our schemes (all three variants) over DCVC (ANFIC) and M-LVC (ANFIC) becomes more significant in terms of PSNR-RGB, while CANF-VC$^-$ and CANF-VC Lite show comparable or better MS-SSIM-RGB results than DCVC (ANFIC). Interestingly, the gap in PSNR-RGB between the more capable HM and the learned coders is still considerable, although the latter outperform HM in terms of MS-SSIM-RGB.

\begin{figure*}[t]
\centering
\resizebox{0.90\textwidth}{!}
{
\Large
\begin{tabular}{c|c|c|c|c}
    Ground Truth & DCVC (ANFIC) & CANF-VC & DCVC-ssim (ANFIC) & CANF-VC-ssim \\
    \includegraphics[width=0.4\textwidth]{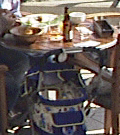}&
    \includegraphics[width=0.4\textwidth]{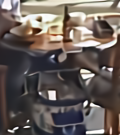}&
    \includegraphics[width=0.4\textwidth]{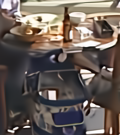}&
    \includegraphics[width=0.4\textwidth]{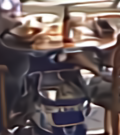}&
    \includegraphics[width=0.4\textwidth]{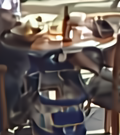}\\
     & PSNR-RGB: $27.71$ dB & PSNR-RGB: $29.00$ dB & MS-SSIM-RGB: $0.952$ & MS-SSIM-RGB: $0.955$ \\
     & $0.0441$ bpp & $0.0396$ bpp & $0.0425$ bpp & $0.0465$ bpp \\
    \includegraphics[width=0.4\textwidth]{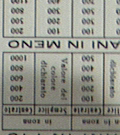}&
    \includegraphics[width=0.4\textwidth]{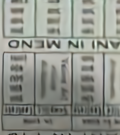}&
    \includegraphics[width=0.4\textwidth]{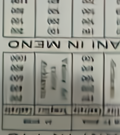}&
    \includegraphics[width=0.4\textwidth]{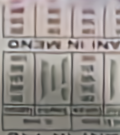}&
    \includegraphics[width=0.4\textwidth]{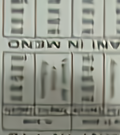}\\
     & PSNR-RGB: $29.29$ dB & PSNR-RGB: $30.23$ dB & MS-SSIM-RGB: $0.946$ & MS-SSIM-RGB: $0.947$\\
     & $0.0373$ bpp & $0.0294$ bpp & $0.0580$ bpp & $0.0573$ bpp\\
\end{tabular}
}
\vspace{-0.3em}
\caption{Subjective quality comparison between CANF-VC and DCVC (ANFIC).}
\vspace{-0.3cm}
\label{fig:visual}
\end{figure*}

\textbf{Subjective Quality Comparison:} Fig.~\ref{fig:visual} presents a subjective comparison between our CANF-VC and DCVC (ANFIC). Both schemes are trained for PSNR-RGB and MS-SSIM-RGB, use ANFIC as the intra-frame coder, and set GOP size to 32. Our CANF-VC is seen to preserve better the shape of the objects and has no color bias, as compared to DCVC (ANFIC).

\vspace{-0.3em}
\subsection{Ablation Experiments}
\vspace{-0.3em}
\label{sec:ablation}
In this section, unless otherwise stated, all the experiments are conducted on UVG dataset~\cite{uvg}, with the BD-rates reported against x265 in veryslow mode.


\textbf{Conditional Inter-frame Coding~vs.~Residual Coding:} To single out the gain of conditional inter-frame coding over residual coding, Table~\ref{tab:study_inter} presents a breakdown analysis in terms of BD-rate savings. In this ablation experiment, the conditional motion coding is disabled and replaced with the motion coder from DVC~\cite{dvclu}. Besides, the residual coding schemes adopt ANFIC~\cite{anfic} for coding the residual frame $x_t-x_c$ as an intra image. The variants with the temporal prior additionally involve the motion-compensated frame $x_c$ in estimating the coding probabilities of the latent code (i.e. $\hat{z}_2$ in Fig.~\ref{fig:canf}). As seen in the table, the conditional inter-frame coding outperforms the residual coding significantly, whether the temporal prior is enabled or not. This suggests that a direct application of ANFIC to residual coding is unable to achieve the same level of gain as our CANF-based inter-frame coding. The temporal prior additionally improves the rate savings of both schemes by 
2.5\% to 4.2\%. 

\textbf{Conditional Motion Coding~vs.~Predictive/Intra Motion Coding:} This ablation experiment addresses the benefits of conditional motion coding. To this end, different competing settings employ the same conditional inter-frame coding, but change the way the flow map $f_t$ is coded. The baseline settings use ANFIC~\cite{anfic} to code $f_t$ as an intra image or the flow map residual $f_t-f_c$ without any condition. For the conditional motion coding, we additionally present results by simply using the previously decoded flow map $\hat{f}_{t-1}$ as the condition. Separate models are trained for each test case. From Table~\ref{tab:study_motion}, our conditional motion coding (i.e. coding $f_t$ based on $f_c$) achieves the best performance. In terms of rate savings, its gain over the two unconditional variants, i.e. coding $f_t$ or $f_t-f_c$ unconditionally, is quite significant. This result corroborates the superiority of our conditional motion coding to predictive motion coding (i.e. coding $f_t-f_c$). As expected, the quality of the condition has a crucial effect on compression performance. The trivial use of the previously decoded flow $\hat{f}_{t-1}$ does not show much gain as compared to unconditional coding. The fact substantiates the effectiveness of our extrapolation network. 

\begin{table}[t!]
    \begin{subtable}[t]{\textwidth}
        \centering
        \tabcolsep=7pt
        \begin{tabular}{cccc}
            \toprule
            Cond. Inter-frame Coding & Residual Coding & Temporal Prior & BD-Rate \\\hline
                                     & $\checkmark$       &                   & -21.8\%    \\
                                     & $\checkmark$       & $\checkmark$      & -24.3\%    \\
            $\checkmark$             &                    &                   & -28.9\%    \\
            $\checkmark$             &                    & $\checkmark$      & -33.1\%    \\
            \bottomrule
        \end{tabular}
        \vspace{-0.2em}
        \caption{}
        \vspace{-0.2em}
        \label{tab:study_inter}
    \end{subtable}
    \begin{subtable}[t]{0.47\textwidth}
        \vspace{-0.2em}
        \centering
        \tabcolsep=3pt
        \begin{tabular}[t]{cccc}
                \toprule
                Input of Motion Coder   & Cond.       & BD-Rate     \\ \hline
                $f_t$                   & -               &    -33.4\%  \\
                $f_t - f_c$             & -               &    -35.3\%  \\
                $f_t$                   &$\hat{f}_{t-1}$  &    -35.2\%  \\ 
                \textcolor{blue}{$f_t$} &\textcolor{blue}{$f_c$}   &    \textcolor{blue}{-42.5\%}  \\
                \bottomrule
            \end{tabular}
        \vspace{-0.2em}
        \caption{}
        \label{tab:study_motion}
    \end{subtable}
    \hspace{0.5em}
    \begin{subtable}[t]{0.47\textwidth}
        \vspace{-0.2em}
        \centering
        \begin{tabular}[t]{cccc}
            \toprule
            Motion Coder  & Inter-frame Coder & BD-Rate \\\hline
            DVC~\cite{dvclu} & 1-step CANF    & -31.4\%  \\
            DVC~\cite{dvclu} & 2-step CANF    & -33.1\%  \\ \hline
            1-step CANF      & 2-step CANF    & -38.1\%  \\
            \textcolor{blue}{2-step CANF}     & \textcolor{blue}{2-step CANF}       & \textcolor{blue}{-42.5\%}  \\
            \bottomrule
        \end{tabular}
        \vspace{-0.2em}
        \caption{}
        \label{tab:study_layer}
     \end{subtable}
     
     \begin{subtable}[t]{\textwidth}
        \vspace{-0.2em}
        \centering
        \tabcolsep=5pt
        \begin{tabular}{ccc}
            \toprule
            Motion Coder      & Inter-frame Coder          & BD-Rate    \\ \hline
            1-step CANF                   & 1-step CANF                   & 35.3\%    \\
            \textcolor{blue}{2-step CANF} & \textcolor{blue}{2-step CANF} & \textcolor{blue}{-42.5\%}   \\
            3-step CANF                   &    3-step CANF                & -31.4\%    \\
            \bottomrule
        \end{tabular}
        \vspace{-0.2em}
        \caption{}
        \label{tab:study_inter_2}
    \end{subtable}

    \vspace{-0.4em}
    \caption{(a) Comparison of conditional inter-frame coding and residual coding under the settings with and without the temporal prior. (b) Comparison of the conditional motion coding, predictive motion coding, and intra motion coding. (c)(d) Comparisons of the conditional motion and inter-frame coders with a varied number of autoencoding transforms. The rows with blue color are our proposed full model.}
    \vspace{-2em}
\end{table}

\textbf{The Number of Autoencoding Transforms:}
\label{sec:layer_compare}
Table~\ref{tab:study_layer} explores the effect of the number of autoencoding transforms on compression performance. The 1-step models are obtained by skipping the autoencoding transform $\{g^{enc}_{\pi_1},g^{dec}_{\pi_1}\}$ in Fig.~\ref{fig:canf}. To have the model size compatible with the 2-step models, the 1-step models have more channels in every autoencoding transform. We first experiment with the conditional inter-frame coding, with the motion coder from DVC~\cite{dvclu}. In this case, the 2-step model improves the rate saving of the 1-step model by 1.7\%. Given the 2-step inter-frame coder, it is further seen that the 2-step motion coder also improves the rate saving of the 1-step motion coder by 4.4\%. This suggests that with a similar model size, the 2-step model is superior to the 1-step model in both inter-frame and motion coding.

Table~\ref{tab:study_inter_2} complements Table~\ref{tab:study_layer} to present results for 1-, 2- and 3-step CANF when applied to both the motion and inter-frame codecs. 3-step CANF extends straightforwardly the 2-step CANF by incorporating one additional autoencoding transform. Despite a larger capacity, the 3-step CANF performs worse than the 2-step CANF and comparably to the 1-step CANF. From Fig.~\ref{fig:canf}, the quantization error introduced to the latent code $z_2$ and the approximation error between $x_c$ (used for decoding) and $y_2$ (generated during encoding) are propagated and accumulated (from top to bottom in Fig.~\ref{fig:canf}) during decoding. The cascading effect, compounded by temporal error propagation, may outweigh the benefits of having more autoencoding transforms.
\section{Conclusion}
\label{sec:conclude}
This work introduces CANF-VC for conditional inter-frame and motion coding. CANF-VC achieves the state-of-the-art video compression performance. Our major findings include: (1) the CANF-based inter-frame coding outperforms residual coding; (2) likewise, our conditional motion coding outperforms predictive motion coding at the cost of additional buffer requirements; (3) the quality of the conditioning variable is critical to compression performance; (4) our 2-step CANF performs better than 1-step CANF, justifying the use of multi-step CANF. 
Lastly, we note that CANF-VC does not use auto-regressive models in inter-frame and motion coding. Its operations are parallelizable. 


\vspace{-0.5em}
\section*{Acknowledgements}
\vspace{-0.5em}

This work was supported by MediaTek, National Center for High-Performance Computing, Taiwan, Ministry of Science and Technology, Taiwan under Grand Application 110-2221-E-A49-065-MY3 and 110-2634-F-A49-006-, and Italian Ministry of University and Research under Grant Application PRIN 2022N25TSZ.

\clearpage
%
%
\bibliographystyle{splncs04}
\bibliography{egbib}
\end{document}


\pagestyle{headings}
\mainmatter
\def\ECCVSubNumber{3904}  

\title{CANF-VC: Conditional Augmented Normalizing Flows for Video Compression \\\emph{Supplementary Materials}} 

\titlerunning{ECCV-22 submission ID \ECCVSubNumber} 
\authorrunning{ECCV-22 submission ID \ECCVSubNumber} 
\author{Anonymous ECCV submission}
\institute{Paper ID \ECCVSubNumber}

\titlerunning{CANF-VC}
%
\author{Yung-Han Ho\inst{1} \and
Chih-Peng Chang\inst{1} \and
Peng-Yu Chen\inst{1} \and
Alessandro Gnutti\inst{2} \and
Wen-Hsiao Peng\inst{1}}
\authorrunning{Ho et al.}
%
\institute{Department of Computer Science, National Yang Ming Chiao Tung University, Hsinchu, Taiwan \email{wpeng@cs.nctu.edu.tw} \and Department of Information Engineering, CNIT, University of Brescia, Brescia, Italy
\email{alessandro.gnutti@unibs.it}}
\maketitle

This supplementary document provides additional materials to assist with the understanding of the performance and design of our CANF-VC. Specifically, it includes:
\begin{itemize}
    \item CANF implementations
    \item Complexity characterization
    \item Rate-distortion curves with GOP 32
    \item Comparison with ELF-VC~\cite{elfvc}
    \item Network details: CANF and motion extrapolation
    \item Temporal prior for motion coding
    \item Training strategy
    \item Subjective quality comparison
    \item Command lines for x265 and HM
\end{itemize}

\section{CANF Implementations}

\begin{figure}[t!]
\begin{center}
    \begin{subfigure}{0.48\linewidth}
        \centering
        \includegraphics[width=\textwidth]{figure/inter_coder.png}
        \caption{}
        \label{fig:canf1}
    \end{subfigure}
    \begin{subfigure}{0.48\linewidth}
        \centering
        \includegraphics[width=\textwidth]{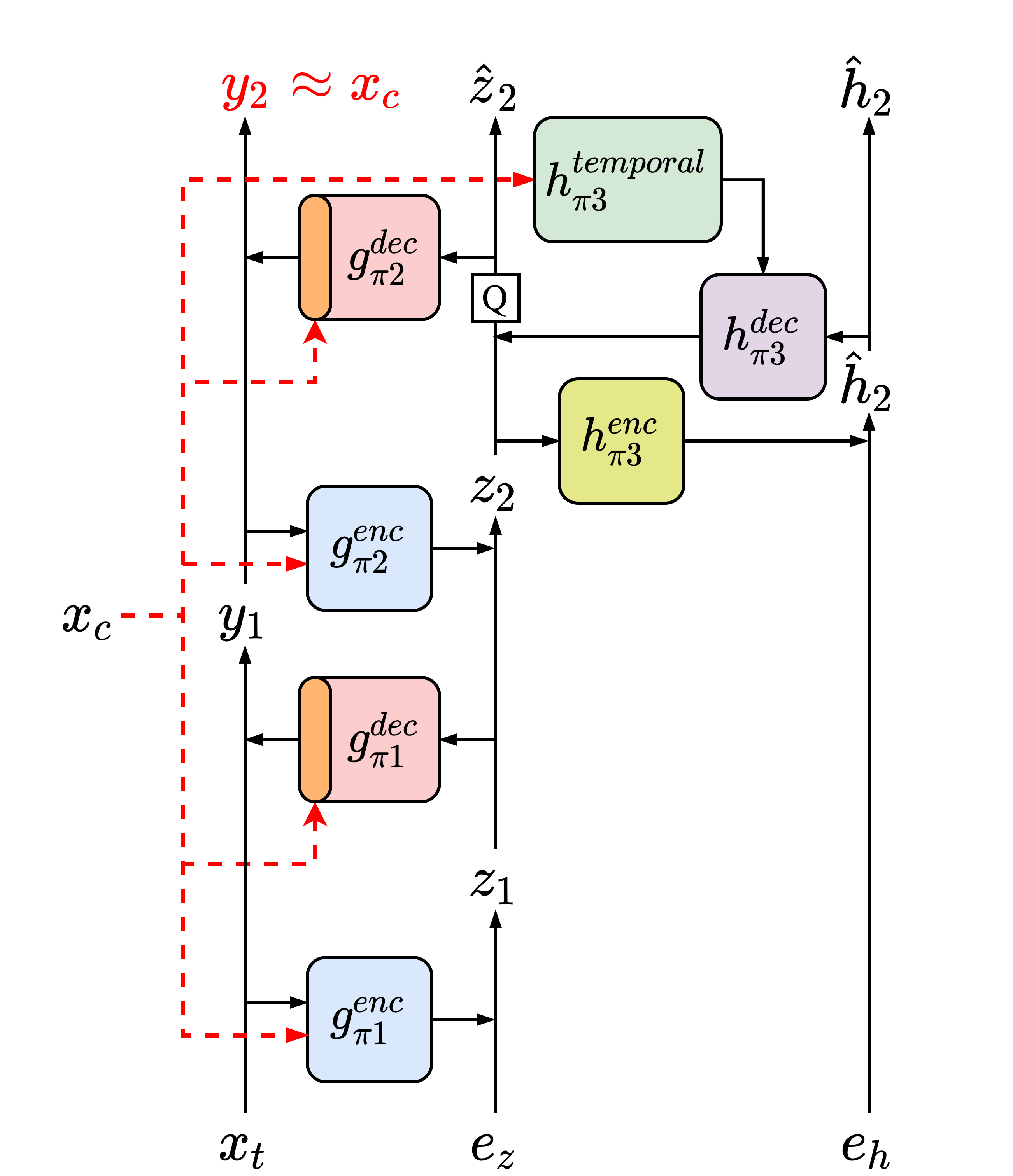}
        \caption{}
        \label{fig:canf2}
    \end{subfigure}

\caption{Illustration of CANF implementations: (a) Decoding transforms w/o $x_c$ and (b) Decoding transforms w/ $x_c$.}
\label{fig:canf_implementation}
\end{center}
\end{figure}

Fig.~\ref{fig:canf_implementation} depicts two possible CANF implementations. Fig.~\ref{fig:canf1} corresponds to the one presented in the main paper. It concatenates the motion-compensated reference frame $x_c$ with the input frame $x_t$ as input to all the encoding transforms. In comparison, the scheme in Fig.~\ref{fig:canf2} additionally accepts $x_c$ as input to all the decoding transforms. The former (decoding transforms w/o $x_c$) can be viewed as a special case of the latter (decoding transforms w/ $x_c$), which utilizes $x_c$ for encoding transforms only. For the implementation of Fig.~\ref{fig:canf2}, the latent code is decoded first to produce 16-channel features having the same spatial resolution as $x_c$. The resulting features are then concatenated with $x_c$ before being processed further by the three convolution layers (the orange part in Fig.~\ref{fig:canf2}) to complete the decoding transform. This implementation (Fig.~\ref{fig:canf2}) has a slightly larger model size than Fig.~\ref{fig:canf1}.

Table~\ref{tab:rd_cond_implement} presents the BD-rate comparison between the two CANF implementations. For experiments, we use the motion coder from DVC~\cite{dvclu}, while the inter-frame coder adopts the two different CANF implementations (Fig.~\ref{fig:canf1} vs. Fig.~\ref{fig:canf2}). It is seen that the more generalized implementation (decoding transforms w/ $x_c$) has comparable performance to our current implementation (decoding transforms w/o $x_c$) on all three datasets. This justifies our choice of decoding transforms w/o $x_c$ because of its comparable performance and simpler design.

\begin{table}[t!]
    \centering
    \caption{BD-rate comparison between two CANF implementations for conditional inter-frame coding. The motion coder is from DVC~\cite{dvclu}. The anchor is x265 in veryslow mode.}
    \label{tab:rd_cond_implement}
    \tabcolsep=7pt
    \resizebox{1.0\textwidth}{!}{
        \begin{tabular}{cccc}
            \toprule
            Implementations              &  UVG~\cite{uvg}  & MCL-JCV~\cite{mcl}   & HEVC-B~\cite{hevc}  \\ 
            \hline
            Decoding transforms w/o $x_c$ & -33.1\%  & -15.3\%   & -35.4\% \\
            Decoding transforms w/ $x_c$  & -35.2\%  & -15.3\%   & -33.9\% \\
            \bottomrule
        \end{tabular}
    }
\end{table}

\section{Complexity Characterization}

Table~\ref{tab:complexity} presents the computational characteristics of different competing methods from the perspectives of multiply-accumulate (MAC) operations, encoding/decoding times for inference, and model sizes. It is to be noted that the prolonged encoding/decoding times of DCVC~\cite{dcvc} are due to the use of an auto-regressive model for entropy coding. Our CANF-VC neither uses an auto-regressive model for motion coding nor uses it for inter-frame coding. Its larger MAC arises from stacking multiple autoencoding transforms. Nevertheless, its relatively short encoding/decoding times suggest that these autoencoding transforms are amenable to parallel computing.
Lastly, we remark that the relatively longer encoding/decoding times of DVC~\cite{dvclu} are due to their software implementation, particularly the entropy coding part. Our CANF-VC follows~\cite{googleiclr18} to quantize the scale parameters from the hyperprior into 64 distinct values, enabling a fast table look-up to derive the probabilities for entropy coding. In contrast, DVC~\cite{dvclu} does not quantize the scale parameters, and needs more time in evaluating higher-precision coding probabilities. 


\begin{table}[t!]
    \centering
    \caption{Complexity characterization in terms of MACs, encoding/decoding times, and model sizes. The MACs and runtimes of DVC~\cite{dvclu} and DCVC~\cite{dcvc} are collected by running the test code released by the respective authors on the same 1080Ti platform. The MACs are evaluated based on encoding a 1080p P-frame, while the encoding/decoding times are averaged over the first 100 P-frames of the Beauty sequence in UVG dataset.}
    \label{tab:complexity}
    \tabcolsep=5pt
    \resizebox{0.9\linewidth}{!}{
        \begin{tabular}{cccc}
            \toprule
            Method                & MACs   & Encoding/Decoding Time  &  Model Size  \\ \hline
            DVC~\cite{dvclu}           & 1725G  & 4.15 s/4.06 s     &  8.5M \\
            DVC\_Pro~\cite{dvcpro}      &   -    &     -/-            &  29M \\
            FVC~\cite{fvc}           &   -    &     -/-    &   26M\\
            DCVC~\cite{dcvc}      & 2268G  & 7.70 s/32.90 s  & 8M\\ \hline
            \textbf{CANF-VC Lite} & 4012G  & 1.38 s/0.98 s  &  15M \\
            \textbf{CANF-VC}      & 5088G  & 1.60 s/1.05 s  & 31M  \\
            \bottomrule
        \end{tabular}
    }
\end{table}


\section{Rate-Distortion Curves with GOP Size 32}

\begin{figure}[t!]
\begin{center}
\begin{subfigure}{0.48\linewidth}
    \centering
    \includegraphics[width=\linewidth]{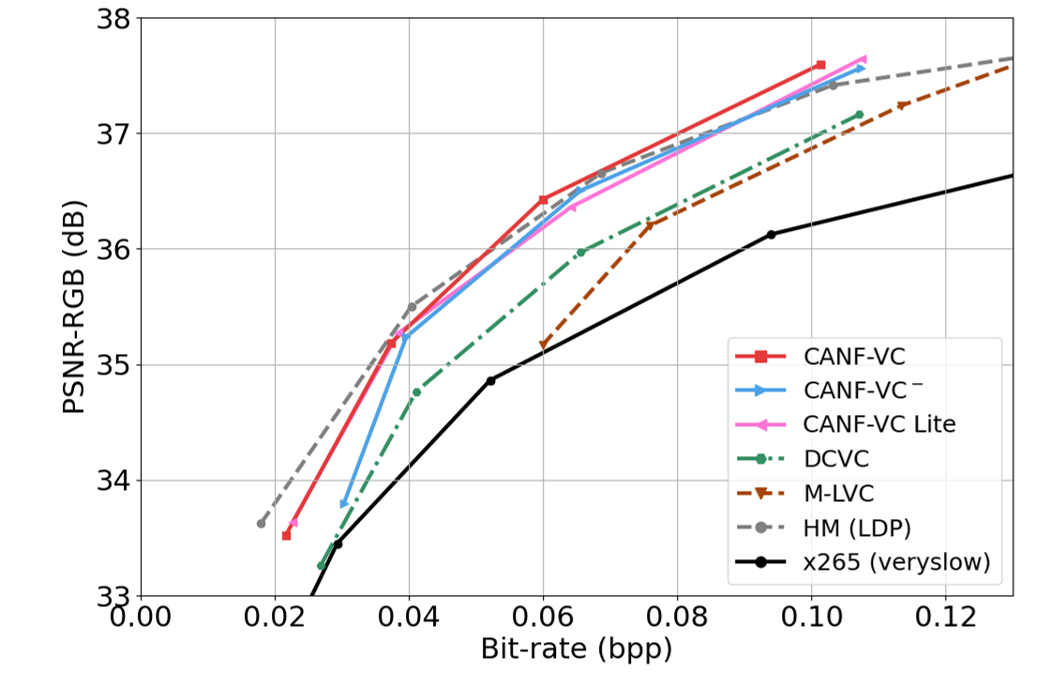} 
    \vspace{-1.5em}
    \caption{UVG, PSNR-RGB}
    \label{fig:uvgPSNR}
\end{subfigure}
\begin{subfigure}{0.48\linewidth}
    \centering
    \includegraphics[width=\linewidth]{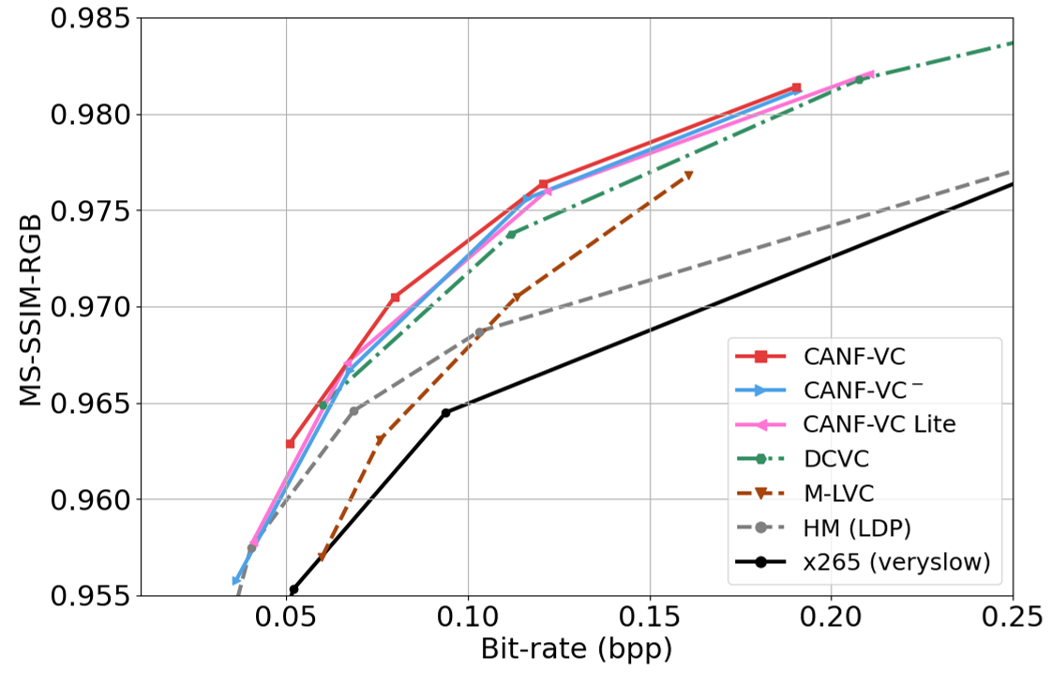}
    \vspace{-1.5em}
    \caption{UVG, MS-SSIM-RGB}
    \label{fig:uvgSSIM}
\end{subfigure}
\begin{subfigure}{0.48\linewidth}
    \centering
    \includegraphics[width=\linewidth]{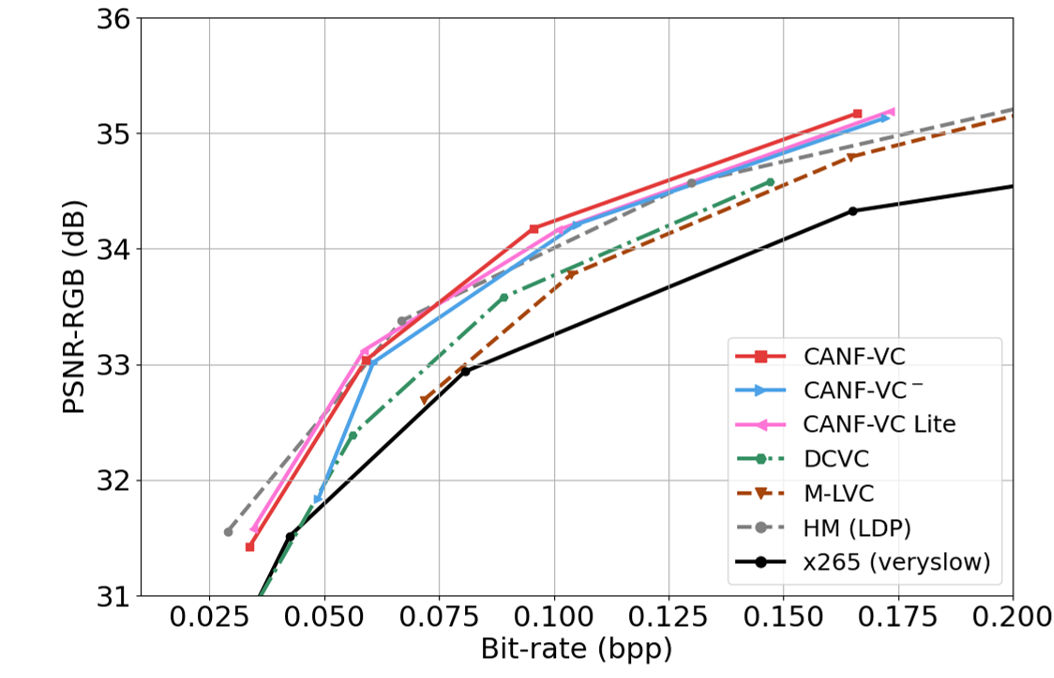}
    \vspace{-1.5em}
    \caption{HEVC Class B, PSNR-RGB}
    \label{fig:hevcPSNR}
\end{subfigure}
\begin{subfigure}{0.48\linewidth}
    \centering
    \includegraphics[width=\linewidth]{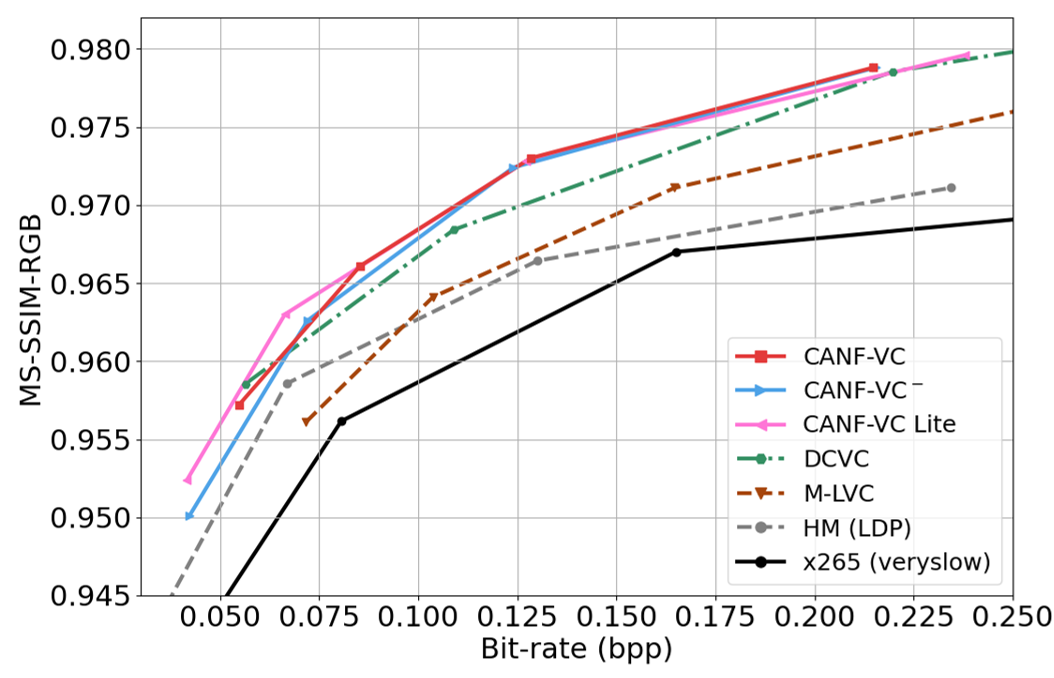}
    \vspace{-1.5em}
    \caption{HEVC Class B, MS-SSIM-RGB}
    \label{fig:hevcSSIM}
\end{subfigure}
\begin{subfigure}{0.48\linewidth}
    \centering
    \includegraphics[width=\linewidth]{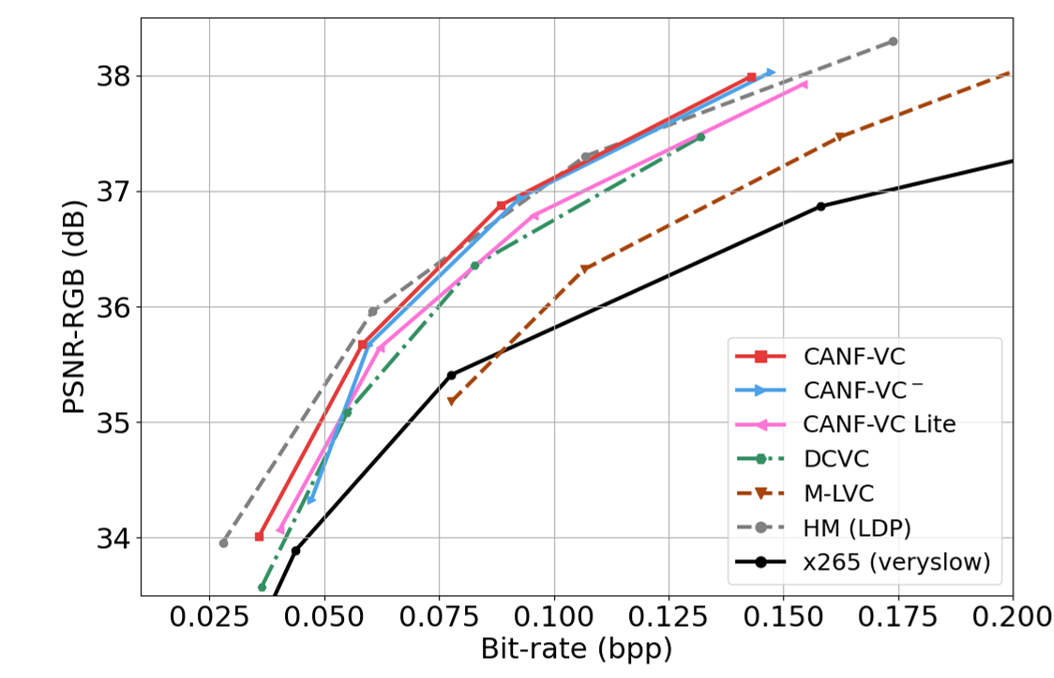}
    \vspace{-1.5em}
    \caption{MCL-JCV, PSNR-RGB}
    \label{fig:mclPSNR}
\end{subfigure}
\begin{subfigure}{0.48\linewidth}
    \centering
    \includegraphics[width=\linewidth]{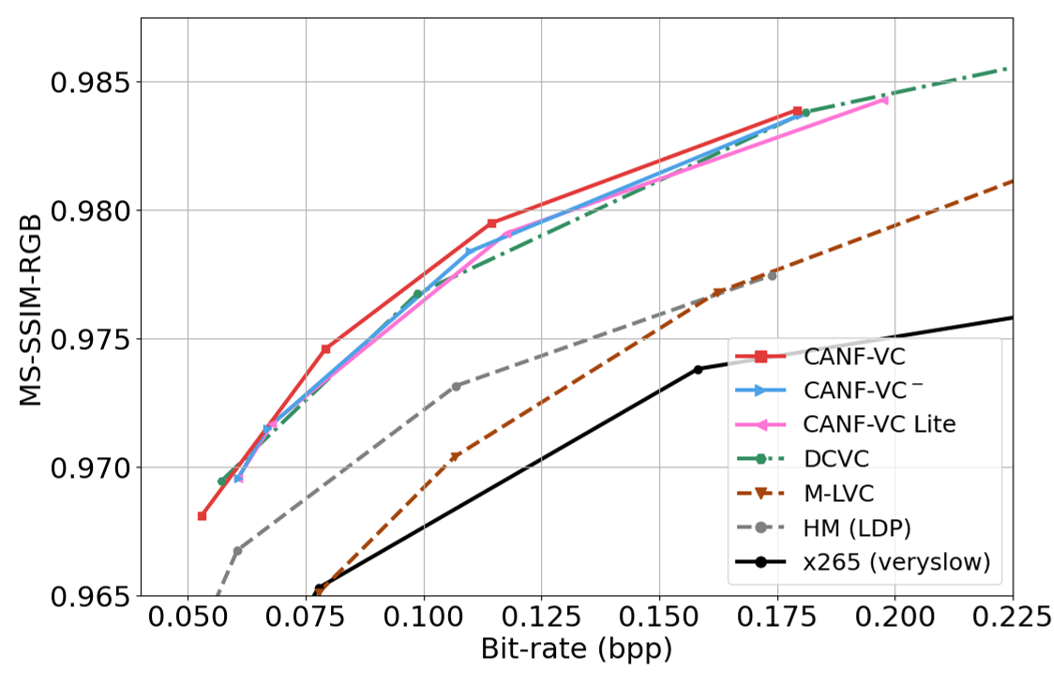}
    \vspace{-1.5em}
    \caption{MCL-JCV, MS-SSIM-RGB}
    \label{fig:mclSSIM}
\end{subfigure}

\caption{Comparison of rate-distortion curves on UVG, HEVC Class B, and MCL-JCV datasets for both PSNR and MS-SSIM. All the competing methods use ANFIC~\cite{anfic} as the intra-frame coder and are evaluated under the same setting, namely, 96-frame encoding with GOP size 32. Results for DCVC~\cite{dcvc} and M-LVC~\cite{mlvc} are produced by their released code.}
\label{fig:RD}
    
\end{center}
\end{figure}

Fig.~\ref{fig:RD} presents rate-distortion curves for HM~\cite{hm}, DCVC~\cite{dcvc}, M-LVC~\cite{mlvc}, and our CANF-VC under GOP size 32 (see Table 2 for their BD-rate figures and Section 4.2 for detailed discussion). Except HM, all the competing methods use ANFIC~\cite{anfic} as the intra-frame coder for a fair comparison. 

In terms of PSNR-RGB, our CANF-VC models outperform DCVC and M-LVC, except for CANF-VC Lite, which performs comparably to DCVC on MCL-JCV dataset. In addition, CANF-VC$^-$ shows worse performance than the other two CANF-VC variants at low rates because it does not include conditional motion coding, which is critical to low-rate compression performance. In comparison with HM, our CANF-VC shows better results at high rates, but worse results at low rates. 

In terms of MS-SSIM-RGB, our CANF-VC models show slightly better results on UVG~\cite{uvg} and HEVC class B~\cite{hevc} than DCVC~\cite{dcvc}, and comparable results on MCL-JCV~\cite{mcl}.

\section{Comparison with ELF-VC~\cite{elfvc}}
Table~\ref{tab:BDrate_gop16} presents separately the BD-rate comparison with ELF-VC~\cite{elfvc} since ELF-VC~\cite{elfvc} adopts a GOP size of 16, which is rarely used by the other competing methods. Note that the results of ELF-VC~\cite{elfvc} are from their paper because its software is unavailable. Moreover, we note that ELF-VC~\cite{elfvc} uses its own intra-frame coder, the details of which are unavailable. Under the same GOP size and in terms of PSNR-RGB, we see that the superiority of our CANF-VC models to ELV-VC~\cite{elfvc} and DCVC (with ANFIC as the intra-frame coder) is obvious. However, ELF-VC~\cite{elfvc} achieves the best MS-SSIM-RGB results among all the competing methods. We remark that this comparison is to provide additional information; a fair comparison would require the software of ELF-VC~\cite{elfvc} and more information about its intra-frame coder.

\begin{table}[t]
    \caption{BD-rate comparison under the same GOP size 16. The anchor is x265 in veryslow mode. Except ELF-VC~\cite{elfvc}, all the competing methods adopt ANFIC~\cite{anfic} as the intra-frame coder.}
    \label{tab:BDrate_gop16}
    \centering
    \tabcolsep=5pt
    \resizebox{\textwidth}{!}{
        \begin{tabular}{lccp{1em}cc}
                \toprule
                \multirow{2}{*} 
                    & \multicolumn{2}{c}{BD-rate (\%) PSNR-RGB} && \multicolumn{2}{c}{BD-rate (\%) MS-SSIM-RGB} \\
                        \cline{2-3} \cline{5-6}
                                        &  UVG   & MCL-JCV   
                                        && UVG   & MCL-JCV  \\ 
                \hline
                DCVC (ANFIC)            &   -19.5 & -7.9     
                                        &&  -47.4 & -46.5   \\ 
                ELF-VC                  &   -30.8 & -11.7                                      
                                        &&  -55.3 & -53.9   \\\hline
                CANF-VC Lite            &   -35.5 & -12.0    
                                        &&  -46.0 & -44.4   \\
                CANF-VC-                &   -34.7 & -13.5    
                                        &&  -45.4 & -43.9   \\
                CANF-VC                 &   -41.0 & -19.9    
                                        &&  -50.3 & -48.9   \\
                \bottomrule
        \end{tabular}
        
    }
\end{table}

\section{Network Details: CANF and Motion Extrapolation}

\begin{figure}[t!]
\centering
\resizebox{0.87\linewidth}{!}{
    \includegraphics[width=\textwidth]{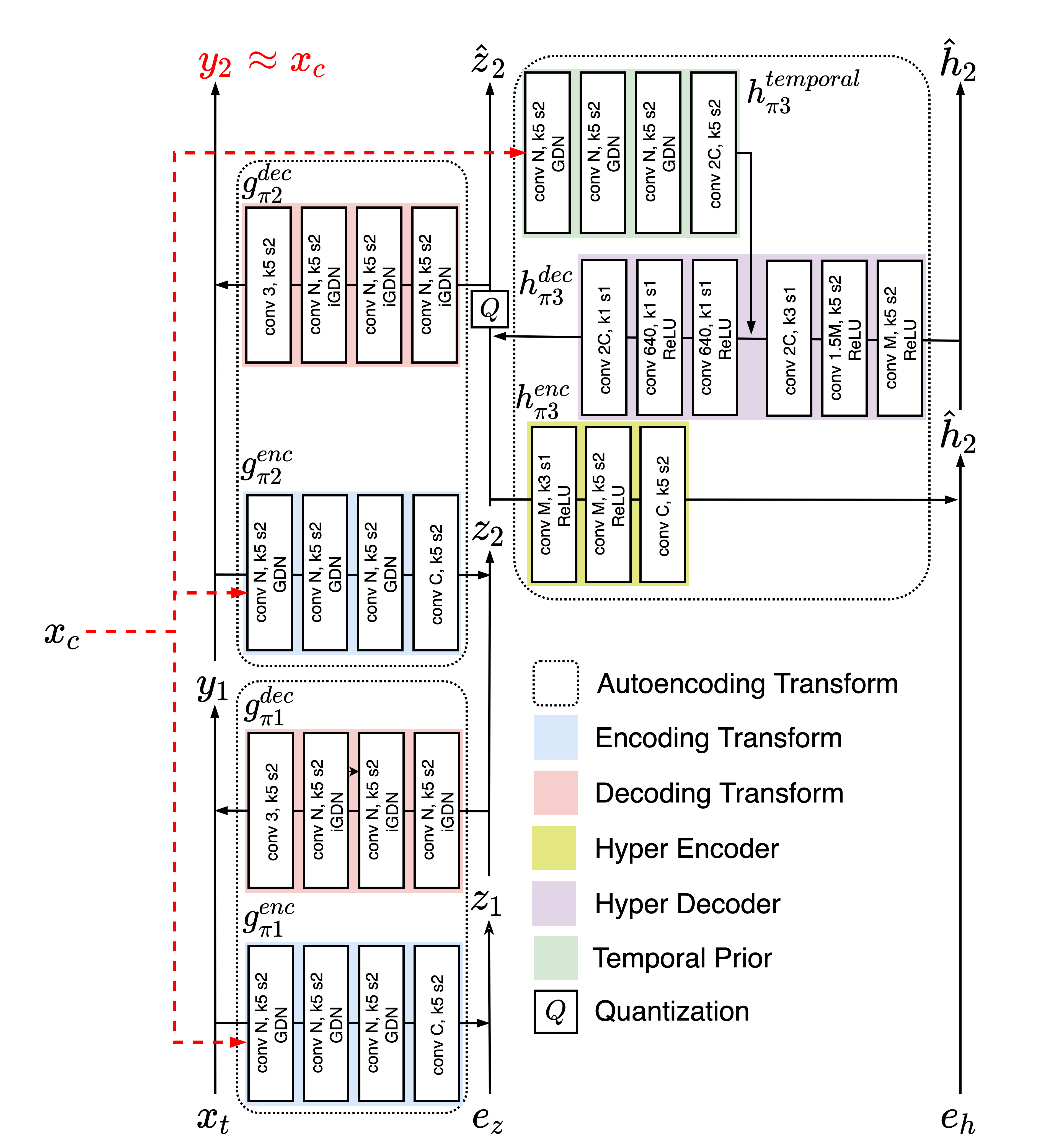}
}
\caption{Network details of our CANF-based coder.}
\label{fig:anfic}
\vspace{-0.5cm}
\end{figure}

Fig.~\ref{fig:anfic} shows the network details of our CANF, where we choose $N=128$ and $C=128$, with $M$ set to 192 for inter-frame coding and 128 for motion coding, respectively. Our CANF-VC Lite adopts $N=72$ and $C=128$, with $M=128$ for both inter-frame and motion coding.

\begin{figure}[t!]
\centering
\includegraphics[width=0.8\linewidth]{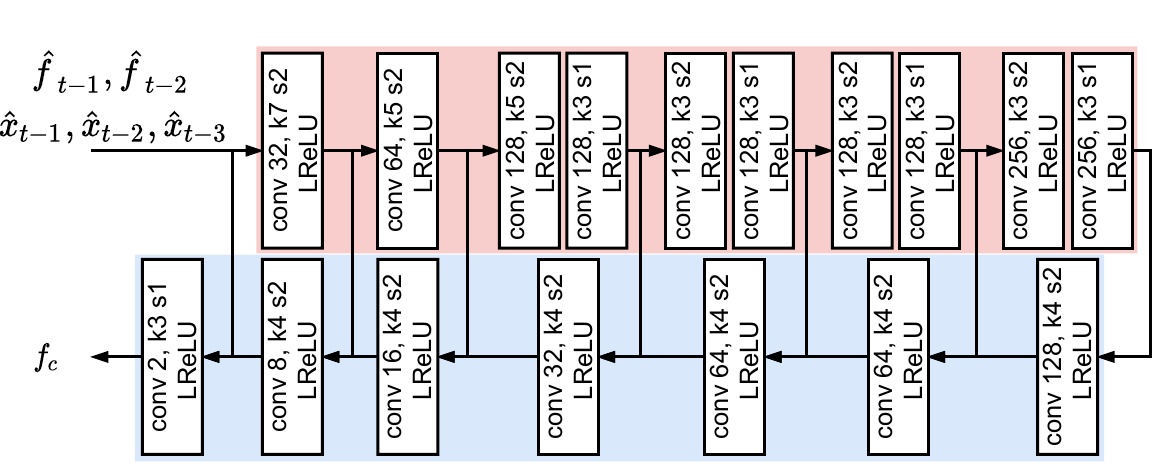}
\caption{Network details of our U-Net-based motion extrapolation network.}
\label{fig:sdc}
\end{figure}

Fig.~\ref{fig:sdc} depicts the network architecture of our U-Net-based motion extrapolation network.

\section{Temporal Prior for Motion Coding}

For conditional motion coding, our current implementation adopts the extrapolated image $warp(\hat{x}_{t-1}; f_c)$ for constructing the temporal prior (Section 3.3 of the main paper). Table~\ref{tab:study_temporal} presents additional results for the case where the predicted flow map $f_c$ is used instead. We see that the former achieves 8\% more rate savings than the latter (i.e. using $f_c$ to construct the temporal prior), which justifies our design choice. The reason may be that the flow map $f_c$ is not as informative as $warp(\hat{x}_{t-1}; f_c)$, which contains more semantic and texture information.

\begin{table}[t!]
    \centering
    \caption{Comparison of different temporal priors for the motion coder.}
    \tabcolsep=7pt
    \label{tab:study_temporal}
    \resizebox{0.8\linewidth}{!}{
        \begin{tabular}{cc}
            \toprule
            Cond. Variable of Temporal Prior & BD-Rate (\%)  \\ \hline
            $warp(\hat{x}_{t-1}; f_c)$  &       -42.5\%      \\
            $f_c$  &      -34.4\%      \\
            \bottomrule
        \end{tabular}
    }
\end{table}

\section{Training Strategy}
Table~\ref{tab:training} summarizes our training steps in three major phases. The first phase uses uncompressed, original frames as inputs to the motion estimation and the motion extrapolation networks, in order to pre-train these networks. We then freeze them until the last three steps. 

In the second (2-frame training) phase, we train the P-frame coder by encoding one P-frame with its reference frame being an uncompressed I-frame. In this phase, the uncompressed frames are used as inputs to the motion estimation and the motion extrapolation networks. We first train the motion coder and the motion compensation network. Subsequently, when the inter-frame coder is involved for training, we fix the motion coder and the motion compensation network for 8 epochs, followed by training jointly the inter-frame and the motion coders for another 5 epochs.

In the third (5-frame training) phase, we use 5 frames (IPPPP) as a basic training unit for forward propagation. However, in updating the P-frame coder, we stop the gradient at each reference frame so that the gradient will not back-propagate through reference frames. In this phase, the previously compressed frames are used as the reference frames (including I-frames) and are input to the motion estimation and the motion extrapolation networks. We train the inter-frame coder, the motion coder, and the motion compensation network for 10 epochs. Lastly, we fine-tune all the networks, including the motion estimation and the motion extrapolation networks, for another 7 epochs with learning rate decay.


\begin{table}[t!]
    \newcommand{\tabincell}[2]{\begin{tabular}{@{}#1@{}}#2\end{tabular}}
    \centering
    \caption{Details of our training strategy.}
    \label{tab:training}
    \tabcolsep=7pt
    \resizebox{\linewidth}{!}{
    \begin{tabular}{c|c|c|c|c}
        \toprule
        Phase              & Training Parts   & 
                           Loss           & lr    & Epochs  
        \\ \hline\hline
        Pre-training       & motion estimation and motion extrapolation networks & 
                           D              &       &        
        \\ \hline
        2-frame (IP) training   & motion coder          &
                           D+$\lambda R$ & 1e-4   & 10
        \\ \hline
                           & motion coder and motion compensation network    &
                           D+$\lambda R$ & 1e-4   & 10
        \\ \hline
                           & Inter-frame coder& 
                           D+$\lambda R$ & 1e-4   & 8
        \\ \hline
                           & \tabincell{c}{motion coder, motion compensation network, \\ and inter-frame coder }& D+$\lambda R$ & 1e-4   & 5
        \\ \hline
        5-frame (IPPPP) training   & \tabincell{c}{motion coder, motion compensation network, \\ and inter-frame coder}&                        D+$\lambda R$ & 1e-4   & 5
        \\ \hline
                                  & \tabincell{c}{motion coder, motion compensation network, \\and inter-frame coder}&                        D+$\lambda R$ & 5e-5   & 5
        \\ \hline
                           & All networks&
                           D+$\lambda R$ & 2.5e-5 & 5
        \\ \hline
                           &All networks&
                           D+$\lambda R$ & 5e-5   & 1
        \\ \hline
                           & All networks&
                           D+$\lambda R$ & 2.5e-5 & 1
        \\ \hline
        \bottomrule
    \end{tabular}
    }
\end{table}

\begin{figure*}[t!]
\centering
\resizebox{0.95\textwidth}{!}
{
\Large
\begin{tabular}{c|c|c|c|c}
    Ground Truth & DCVC (ANFIC) & CANF-VC & DCVC-ssim (ANFIC) & CANF-VC-ssim \\
    \includegraphics[width=0.4\textwidth]{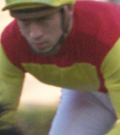}&
    \includegraphics[width=0.4\textwidth]{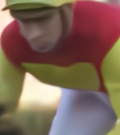}&
    \includegraphics[width=0.4\textwidth]{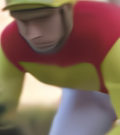}&
    \includegraphics[width=0.4\textwidth]{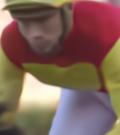}&
    \includegraphics[width=0.4\textwidth]{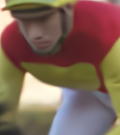}\\
     & PSNR-RGB: $33.84$ dB & PSNR-RGB: $34.50$ dB & MS-SSIM-RGB: $0.967$ & MS-SSIM-RGB: $0.966$ \\
     & $0.0184$ bpp & $0.0109$ bpp & $0.0336$ bpp & $0.0271$ bpp \\
    \includegraphics[width=0.4\textwidth]{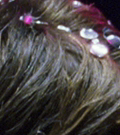}&
    \includegraphics[width=0.4\textwidth]{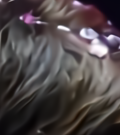}&
    \includegraphics[width=0.4\textwidth]{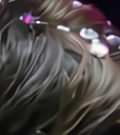}&
    \includegraphics[width=0.4\textwidth]{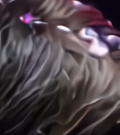}&
    \includegraphics[width=0.4\textwidth]{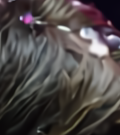}\\
     & PSNR-RGB: $33.84$ dB & PSNR-RGB: $34.19$ dB & MS-SSIM-RGB: $0.956$ & MS-SSIM-RGB: $0.955$ \\
     & $0.0122$ bpp & $0.0090$ bpp & $0.0261$ bpp & $0.0211$ bpp \\
    \includegraphics[width=0.4\textwidth]{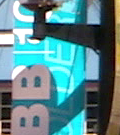}&
    \includegraphics[width=0.4\textwidth]{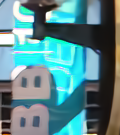}&
    \includegraphics[width=0.4\textwidth]{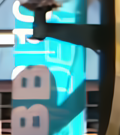}&
    \includegraphics[width=0.4\textwidth]{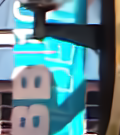}&
    \includegraphics[width=0.4\textwidth]{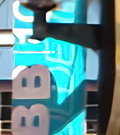}\\
     & PSNR-RGB: $28.09$ dB & PSNR-RGB: $29.02$ dB & MS-SSIM-RGB: $0.957$ & MS-SSIM-RGB: $0.959$ \\
     & $0.0697$ bpp & $0.0638$ bpp & $0.0739$ bpp & $0.0704$ bpp \\
    \includegraphics[width=0.4\textwidth]{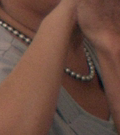}&
    \includegraphics[width=0.4\textwidth]{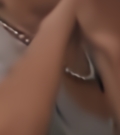}&
    \includegraphics[width=0.4\textwidth]{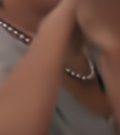}&
    \includegraphics[width=0.4\textwidth]{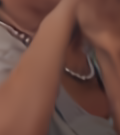}&
    \includegraphics[width=0.4\textwidth]{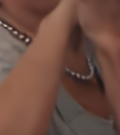}\\
     & PSNR-RGB: $32.68$ dB & PSNR-RGB: $33.26$ dB & MS-SSIM-RGB: $0.972$ & MS-SSIM-RGB: $0.970$ \\
     & $0.0343$ bpp & $0.0267$ bpp & $0.0506$ bpp & $0.0390$ bpp \\
    \includegraphics[width=0.4\textwidth]{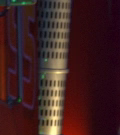}&
    \includegraphics[width=0.4\textwidth]{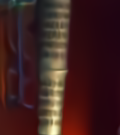}&
    \includegraphics[width=0.4\textwidth]{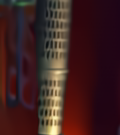}&
    \includegraphics[width=0.4\textwidth]{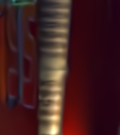}&
    \includegraphics[width=0.4\textwidth]{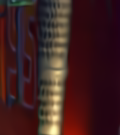}\\
     & PSNR-RGB: $33.96$ dB & PSNR-RGB: $35.50$ dB & MS-SSIM-RGB: $0.979$ & MS-SSIM-RGB: $0.981$ \\
     & $0.0265$ bpp & $0.0232$ bpp & $0.0348$ bpp & $0.0369$ bpp \\
\end{tabular}
}
\vspace{-1em}
\caption{Subjective quality comparison between CANF-VC and DCVC (ANFIC). The suffix "-ssim" indicates that the models are trained with MS-SSIM.}
\vspace{-0.3cm}
\label{fig:visual}
\end{figure*}
\section{Subjective Quality Comparison}

Fig.~\ref{fig:visual} provides more subjective quality comparisons between CANF-VC and DCVC (ANFIC). Results are provided for models trained with PSNR and MS-SSIM. It is seen that our CANF-VC better preserves the shape of the objects than DCVC (ANFIC) (cf. the face of the jockey in the first row, the hair in the second row, the letters "DE" in the third row, the necklace in the forth row, and the textured pattern in the last row).

\section{Command Lines for X265 and HM}

Following~\cite{dcvc}, we use FFmpeg to generate the compressed videos through x265 with veryslow mode. Given an uncompressed video ``input.yuv'' of size W $\times$ H, the command line for x265 encoding is as follows: \emph{ffmpeg -pix fmt yuv420p -s W $\times$H -r FR -i input.yuv -vframes N -c:v libx265 -preset veryslow -tune zerolatency -x265-params “qp=Q:keyint=GOP:verbose=1” output.mkv}, where FR, N, Q, GOP represent the frame rate, the number of frames to be encoded, the quantization parameter and the GOP size, respectively. Q is set to 19, 22, 27, 32, 37. For the common test protocol, we choose GOP to be 10 for HEVC Class B and 12 for the other datasets.

For the encoding with HM, given an uncompressed video ``input.yuv'' of size W $\times$ H, we use the \emph{encoder\_lowdelay\_P\_main.cfg} configuration file~\cite{hm} with the following parameters: InputFile=input.yuv, FrameRate=FR, SourceWidth=W, SourceHeight=H, FramesToBeEncoded=N, IntraPeriod=32, GOPSize=8, DecodingRefreshType=2, and QP=Q, where FR, N, Q represent the frame rate, the number of frames to be encoded, and the quantization parameter, respectively. Q is set to 17, 22, 24, 27, 32.

%
%
\bibliographystyle{splncs04}
\bibliography{egbib}